\documentclass{article}

\PassOptionsToPackage{numbers, compress}{natbib}

\usepackage[final]{neurips_data_2022}

\usepackage[utf8]{inputenc} 
\usepackage[T1]{fontenc}    
\usepackage{hyperref}       
\usepackage{url}            
\usepackage{booktabs}       
\usepackage{amsmath}
\usepackage{amsfonts}       
\usepackage{nicefrac}       
\usepackage{microtype}      
\usepackage{xcolor}         
\usepackage{graphicx}
\usepackage{caption}
\usepackage{wrapfig}
\usepackage{subcaption}
\usepackage[export]{adjustbox}
\usepackage[frozencache,cachedir=.]{minted}
\usepackage{listings}
\definecolor{darkgreen}{rgb}{0, 0.465, 0}
\definecolor{lightgray}{rgb}{.9,.9,.9}
\definecolor{darkgray}{rgb}{.4,.4,.4}
\definecolor{purple}{rgb}{0.65, 0.12, 0.82}
\lstdefinelanguage{yaml}{
  keywords={Player, AvatarObject, Termination, Win, eq, Objects, name, Observers, MapCharacter, Sprite2D, Image, Z, TilingMode, Actions, Object, Dst, Src, Commands, Reward, cascade, Behaviours, mov, remove, Levels, TileSize, BackgroundTile,Environment},
  keywordstyle=\color{darkgreen}\bfseries,
  keywords=[2]{},
  keywordstyle=[2]\color{purple}\bfseries,
  identifierstyle=\color{black},
  sensitive=false,
  comment=[l]{\#},
  morecomment=[s]{/*}{*/},
  commentstyle=\color{gray}\ttfamily,
  stringstyle=\color{red}\ttfamily,
  morestring=[b]',
  morestring=[b]"
}
\title{GriddlyJS: A Web IDE for Reinforcement Learning}

\author{%
  Christopher Bamford\thanks{Work done at Meta AI.} \\
  Queen Mary University\\
  \texttt{c.d.j.bamford@qmul.ac.uk} \\
  \And
  Minqi Jiang\\
  Meta AI \& UCL\\
  \texttt{msj@meta.com}
  \And
  Mikayel Samvelyan\\
    Meta AI \& UCL\\
  \texttt{samvelyan@meta.com} 
  \And
  Tim Rocktäschel\\
  UCL \\
  \texttt{tim.rocktaschel@ucl.ac.uk}
}

\begin{document}

\maketitle

\begin{abstract}
Progress in reinforcement learning (RL) research is often driven by the design of new, challenging environments---a costly undertaking requiring skills orthogonal to that of a typical machine learning researcher. The complexity of environment development has only increased with the rise of procedural-content generation (PCG) as the prevailing paradigm for producing varied environments capable of testing the robustness and generalization of RL agents. Moreover, existing environments often require complex build processes, making reproducing results difficult. 
To address these issues, we introduce GriddlyJS, a web-based Integrated Development Environment (IDE) based on the Griddly engine. GriddlyJS allows researchers to visually design and debug arbitrary, complex PCG grid-world environments using a convenient graphical interface, as well as visualize, evaluate, and record the performance of trained agent models. By connecting the RL workflow to the advanced functionality enabled by modern web standards, GriddlyJS allows publishing interactive agent-environment demos that reproduce experimental results directly to the web. To demonstrate the versatility of GriddlyJS, we use it to quickly develop a complex compositional puzzle-solving environment alongside arbitrary human-designed environment configurations and their solutions for use in automatic curriculum learning and offline RL.
The GriddlyJS IDE is open source and freely available at \url{https://griddly.ai}.
\end{abstract}

\section{Introduction}
\label{sec:intro}

Deep reinforcement learning (RL) has seen rapid progress over the past decade, with recent methods producing policies that match or exceed human experts in tasks ranging from games like Go and Chess \citep{alphago,silver2017mastering} to advanced scientific applications such as plasma control \citep{degrave2022magnetic} and chip design \citep{mirhoseini2020chip}. 
Concurrent with this progress is the proliferation of increasingly challenging RL environments that serve as the necessary experimental substrates for such research breakthroughs \citep{cobbe_2019a, nle}.
In RL, environments testing specific agent capabilities serve as crucial measuring sticks against which progress is assessed, playing an analogous role to  benchmarks such as MNIST \citep{mnist}, ImageNet \citep{imagenet}, and GLUE \citep{glue} in supervised learning. For example, the Procgen Benchmark tests for generalization \citep{cobbe_2019a}, and D4RL, for offline RL performance \citep{fu2020d4rl}.
Developing RL methods for new problem settings requires new environments specifically embodying such settings. Environment development thus plays a crucial role in research progress. Moreover, the availability of a large, diverse selection of environments mitigates the risk of overfitting our methods to a small set of tasks \citep{cobbe_2019a, nle}.

\begin{figure}[t!]
    \centering
    \includegraphics[width=1\textwidth, center]{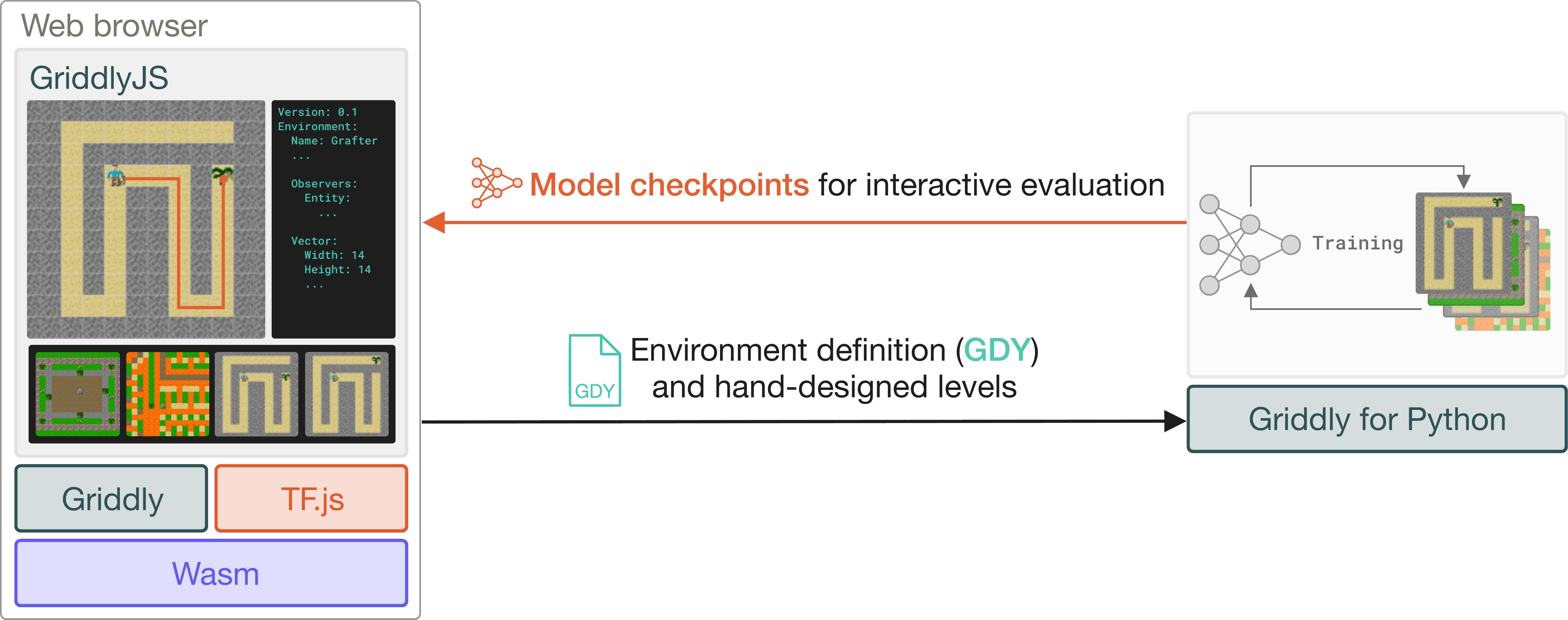}
    \caption{\small{An overview of the human-in-the-loop environment development workflow enabled by \mbox{GriddlyJS}, built on top of the Griddly engine in Wasm and Tensorflow.js (TF.js). Environments and custom designed levels can be loaded into Griddly for Python for training, and model checkpoints can be directly loaded into GriddlyJS for visual evaluation.}}
    \label{fig:griddly_js_system}
    \vspace{-7mm}
\end{figure}

Unfortunately, developing new environments that provide the necessary challenges for RL methods is a costly process, requiring deep expertise in software engineering, high performance computing (for efficient distributed training), and game design. This skillset is closer to that of a videogame developer than a typical machine learning researcher. 
Recently, procedural content generation (PCG) has emerged as the standard paradigm for developing environments that can vary throughout training, enabling the study of systematic generalization and robustness in RL \citep{risi2020increasing, nle, cobbe_2019a, gym_minigrid, juliani2019obstacle, samvelyanMiniHackPlanetSandbox2021}. The more complex programming logic entailed by PCG algorithms, i.e. creating probabilistic programs that specify distributions over environment configurations, adds considerable engineering overhead to the creation of new RL environments \citep{azadScenic4RLProgrammaticModeling2021}. 
As environments grow in number and complexity, researchers pay the additional cost in managing their associated build pipelines and dependencies---an often time-consuming task. This overhead can make reproducing results a difficult ordeal, even when pre-trained model checkpoints are provided. 
Moreover, most environments do not offer any tooling for visualizing, evaluating, or recording agent trajectories. Currently, writing code to support these activities demands a large time commitment from researchers, slowing down the pace of research progress and adding further obstacles to debugging and assessing existing methods.

To address these challenges, we introduce GriddlyJS, a web-based integrated development environment (IDE) based on a WebAssembly (Wasm) version of the Griddly engine \citep{bamford_2021}. GriddlyJS provides a simple interface for developing and testing arbitrary, procedurally-generated grid-world environments in Griddly using a visual editor and domain-specific language based on YAML, with support for highly complex game mechanics and environment generation logic. The visual editor allows rapid design of new levels (i.e. variations of an environment) via a simple point-and-click interface. Environments can be tested via interactive control of the agent directly inside the IDE.

GriddlyJS produces Griddly game description YAML files (GDY), which define environments and custom levels. GDY files can be loaded directly into Griddly for Python, producing a Gym-compatible environment. In addition, any agent model can be loaded into the GriddlyJS IDE, once easily converted to the TensorFlow.js \citep[TF.js;][]{tfjs} format, allowing visualizing, evaluating, or recording performance. 
The integrated development and visualization provided by GriddlyJS enables a whole new mode of closed-loop development of RL environments, in which the researcher can rapidly iterate on environment design based on the behavior of the agent. This allows designing environments that specifically break state-of-the-art RL methods, thereby more quickly pushing the field forward.

Importantly, GriddlyJS runs the Griddly runtime directly inside the browser. The environment simulation is rendered as a modular web component based on the React library \citep{react}. This design allows researchers to easily publish fully interactive demos of pre-trained agent models in Griddly environments directly on the web, as embedded components in a webpage. This simplified sharing of agent-environment demos allows new RL results to be rapidly and comprehensively verified. It also enables the collection of novel, challenging environments designed by humans, providing similar benefits for improving robustness as recent adversarial human-in-the-loop training strategies in the supervised question and answering (QA) domain \citep{kiela2021dynabench}---all without building and installing any additional software.



In the remainder of this paper, we provide a detailed description of the GriddlyJS IDE and highlight how it addresses existing environment-centric bottlenecks to RL research. As an exemplary use-case, we then demonstrate how GriddlyJS can be used to quickly design a new environment that require solving complex, compositional puzzles, as well as efficiently generate a new offline RL dataset based on human play-throughs in this environment. Additionally, we show how GriddlyJS can be used to rapidly produce human-designed levels, which we demonstrate empirically to be a promising approach for producing more difficult levels for RL agents than procedural content generation.

\section{Background}

\subsection{Reinforcement Learning} \label{sec:rl}
GriddlyJS streamlines the design of reinforcement learning (RL) environments. Such environments can generally be represented as a partially-observable Markov Decision Process \citep[POMDP;][]{sondik1978optimal}, defined as a tuple $\mathcal{M} = (S, A, O, \Omega, \mathcal{T}, R, \gamma, \rho)$, where $S$ is the state space, $A$ is the action space, $O$ is the observation space, $\Omega:S \rightarrow O$ is the observation (or emission) function, $\mathcal{T}:S\times A \rightarrow S$ is the transition function, $R:S \rightarrow \mathbb{R}$ is the reward function, $\gamma$ is the discount factor, and $\rho$ is the distribution over initial states. At each time $t$, the RL agent takes an action $a_t$ according to its policy $\pi(a_t|o_t)$, where $o_t \sim \Omega(s_t)$, and the environment then transitions its state to $s_{t+1} \sim \mathcal{T}(s_t, a_t)$, producing a reward $r_t = R(s_{t+1})$ for the agent. The RL problem seeks to maximize the expected return, that is, the sum of future discounted rewards over the initial state distribution $\rho$. Given the policy $\pi$ is parameterized by $\theta$, this goal is captured by the objective $J(\theta) = \mathbb{E}_{\pi} \left[ \sum_{t} \gamma^t r_t \right]$, where the expectation is over state-action trajectories arising from following $\pi$ in $\mathcal{M}$ with $s_0 \sim \rho$.

\subsection{Procedural Content Generation}
Recent focus in RL research has shifted from singleton environments, which take exactly the same form across each training episode, to environments making use of procedural content generation (PCG) that can vary specific aspects throughout training. Policies trained on singleton environments, like those in the Atari Learning Environment Benchmark \citep{bellemare2013arcade} are overfit to the single environment instance, failing when even small changes are applied to the original training environment \citep{farebrother2018generalization}. In contrast, PCG environments produce an endless series of environment configurations, also called \emph{levels}, by modifying specific environment aspects algorithmically per episode, e.g. the appearance, layout, or even specific transition dynamics. PCG environments can typically be deterministically reset to specific levels by conditioning the underlying PCG algorithm on a random seed. 

PCG environments allow us to evaluate the robustness and systematic generalization of RL agents. The evaluation protocol follows that in supervised learning, whereby the agent is trained on a fixed number of training levels and tested on held-out levels. Environment levels typically share a common dynamical structure, including a shared state and action space, allowing for learning transferable behaviors across levels. As a result, providing the agent with more levels at training typically leads to stronger test-time performance \citep{cobbe_2019a}. A separate, commonly used training method called \emph{domain randomization} \citep[DR;][]{tobin2017domain} simply resets the environment to a random level per episode, based on the underlying PCG algorithm. DR can produce robust policies and can be viewed as a form of data augmentation over the space of environment configurations. Adaptive curricula methods that more actively generate levels have been shown to further improve robustness in many PCG domains~\citep{portelas2020teacher,dennisEmergentComplexityZeroshot2021, jiang2021prioritized}. Most real-world domains exhibit considerable diversity, making PCG an important paradigm for producing environments better equipped for sim2real transfer.

\subsection{The Case for Grid Worlds}
\label{subsec:gridworlds_background}
Grid worlds are environments corresponding to MDPs with discrete actions and states that can be represented as a 3D tensor. Note that while the state is constrained to be a tensor with dimensions $M \times N \times K$, where  $M,N,K \in \mathbb{Z}^+$, the actual observations seen by the agent may be rendered differently, e.g. in pixels or as a partial observation. Typically, in 2D grid worlds, each position in the grid, or $\emph{tile}$, corresponds to an entity, e.g. the main agent, a door, a wall, or an enemy unit. The entity type is then encoded according to a vector in $R^{K}$. By constraining the state and action space to simpler, discrete representations, grid worlds drastically cut down the computational cost of training RL agents without sacrificing the ability to study the core challenges of RL, such as exploration, planning, generalization, and multi-agent interactions. 

Indeed, many of the most challenging RL environments, largely unsolved by even the latest state-of-the-art methods, are PCG grid worlds. For example, Box-World \citep{zambaldi2018relational} and RTFM \citep{zhong2020rtfm} are difficult grid worlds that require agents to perform compositional generalization; many games of strategy requiring efficient planning such as Go, Chess, and Shogi may be formulated as grid worlds; and many popular exploration benchmarks, such as MiniHack \citep{samvelyanMiniHackPlanetSandbox2021} and MiniGrid \citep{gym_minigrid} take the form of grid worlds. A particularly notable grid world is the NetHack Learning Environment \citep[NLE;][]{nle}, on which symbolic bots currently still outperform state-of-the-art deep RL agents \citep{hambro2022insights}. NLE pushes existing RL methods to their limits, requiring the ability to solve hard exploration tasks and strong systematic generalization, all in a partially-observable MDP with extremely long episode lengths. Crafter is a recent grid world that features an open-world environment in which the agent must learn dozens of skills to survive, and where the strongest model-based RL methods are not yet able to match human performance \citep{crafter}. To demonstrate the potential of GriddlyJS, we use it to create a Griddly-based Crafter in Section~\ref{sec:poc}.

Their common grid structure and discrete action space allows for grid worlds to be effectively parameterized in a generic specification. GriddlyJS takes advantage of such a specification to enable the mass production of diverse PCG grid worlds encompassing arbitrary game mechanics. Thus, while GriddlyJS is limited to grid worlds, we do not see this as significantly limiting the range of fundamental research that it can help enable. Still, it is important to acknowlege that grid worlds can not provide an appropriate environment for all RL research. In particular, grid worlds cannot directly represent MDPs featuring continuous state spaces, including many environments used in robotics research such as MuJoCo~\citep{mujoco} and DeepMind Control Suite~\citep{dmc}. Nevertheless, as we previously argue, grid worlds capture the fundamental challenges of RL, making them an ideal testbed for benchmarking new algorithmic advances. Further, many application domains can directly be modeled as grid worlds or quantized as such, e.g. many spatial navigation problems, video games like NetHack~\citep{nle}, combinatorial optimization problems like chip design~\citep{mirhoseini2021graph}, and generally any MDP with discrete state and action spaces.

\subsection{Griddly}
\begin{figure}[t!]
  \centering
  \begin{subfigure}{0.4\linewidth}
    \centering
    \includegraphics[width=\textwidth]{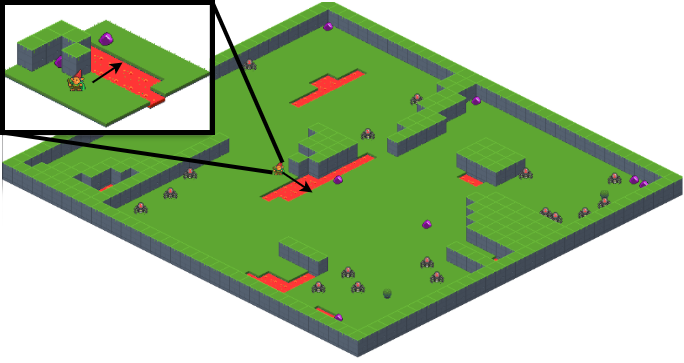}
  \end{subfigure}
  \begin{subfigure}{0.5\linewidth}
    \centering
    \includegraphics[width=\textwidth]{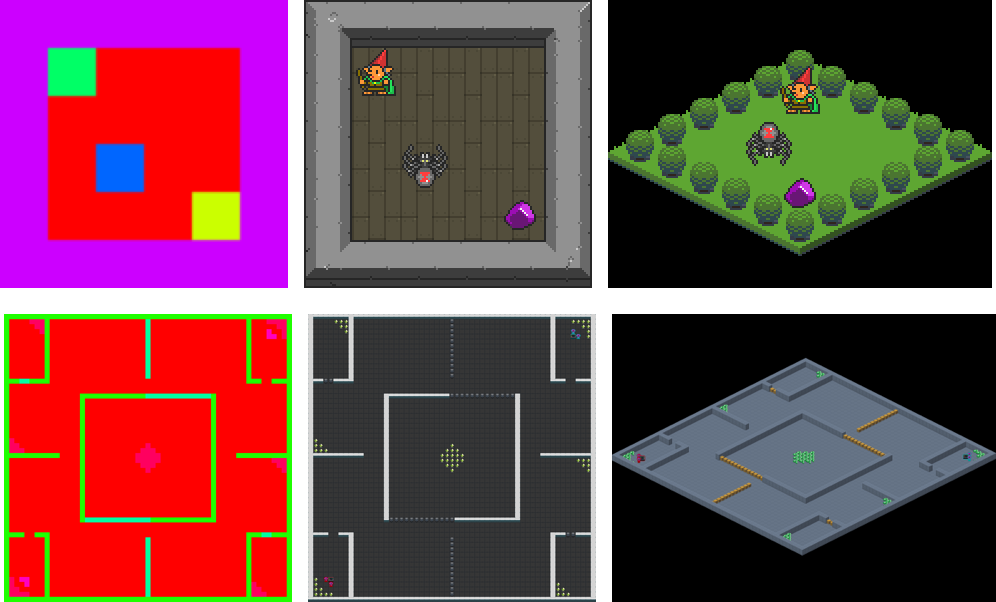}  
  \end{subfigure}%
  \caption{\small{Visualization of some observation spaces supported in Griddly. The left figure shows an isometric, global view of an environment with an example of a local observation. On the right, each row shows the same environment state rendered under three different observers: \texttt{Vector}, \texttt{Sprite2D}, and \texttt{Isometric}}.} 
  \label{fig:observers}
\end{figure}

Griddly\footnote{Documentation, tutorials, examples and API reference can be found on the Griddly documentation website \url{https://griddly.readthedocs.io}} \cite{bamford_2021} is a game engine designed for the fast and flexible creation of grid-world environments for RL, with support for both single and multi-agent environments. Griddly simplifies the implementation of environments with complex game mechanics, greatly improving research productivity. It allows the underlying MDP to be defined in terms of simple declarative rules, using a domain-specific language (DSL) based on Yet Another Markup Language (YAML). This is a similar approach to GVGAI \citep{perezliebana_2018}, MiniHack \citep{samvelyanMiniHackPlanetSandbox2021} and Scenic4RL \citep{azadScenic4RLProgrammaticModeling2021}, where a DSL language is used to define environment mechanics. Griddly's DSL is designed to be low level in terms of interactions between defined objects, but does not go as far to allow the user to define physical models unlike Scenic4RL. This is similar in regards to GVGAI and MiniHack as they are both grid-world games. Griddly's DSL contains higher level functions such as A* search and proximity sensing, which can be configured to build higher-level behaviours for NPCs. MiniHack's DSL gives access to all of the objects within the base NetHack game, and allows them to be configured at a high level using modifiers such as "hostile" or "peaceful". The choice of using YAML is also more flexible, as YAML is a common DSL with supporting libraries in many different languages. This allows the generation and manipulation of GDY files without requiring the construction of parsers or serializers. Integrations of higher level tooling such as GriddlyJS are made possible due to this. Griddly game description YAML files share a similar structure composed of three main sections: \texttt{Environment}, \texttt{Actions}, and \texttt{Objects}. \texttt{Objects} describe the entities within the environment, declare local variables and initialization procedures. \texttt{Actions} define local transition rules determining the evolution of the environment state and any rewards that are received by the agent. Such \texttt{Actions} are distinct from the MDP actions and can be thought of as local transition functions, whose arguments are game entities (and their internal states) and the actions taken by the agent. The \texttt{Environment} section defines the environment description, e.g. its name, as well as the MDP observation and action spaces. For all environments, Griddly uses the canonical action space layout from \cite{bamfordGeneralisingDiscreteAction2021a}, with support for invalid action masking. In Griddly, {\em observers} are functions that determine how any given environment state is rendered into an observation. Several different types of observers are supported, including those shown in Figure \ref{fig:observers}. Griddly additionally supports lightweight observers based on default 2D block sprites, ASCII, and semantic entity states (as defined in the GDY). Observers are highly customizable, with options for partial observability, rotation, cropping, and inclusion of global or internal, entity-level variables. We provide a simple example Griddly environment implementation in Appendix~\ref{sec:gdy_overview}.

\section{GriddlyJS} \label{sec:GriddlyJS}

\begin{figure}[t!]
    \centering
    \includegraphics[width=\textwidth, center]{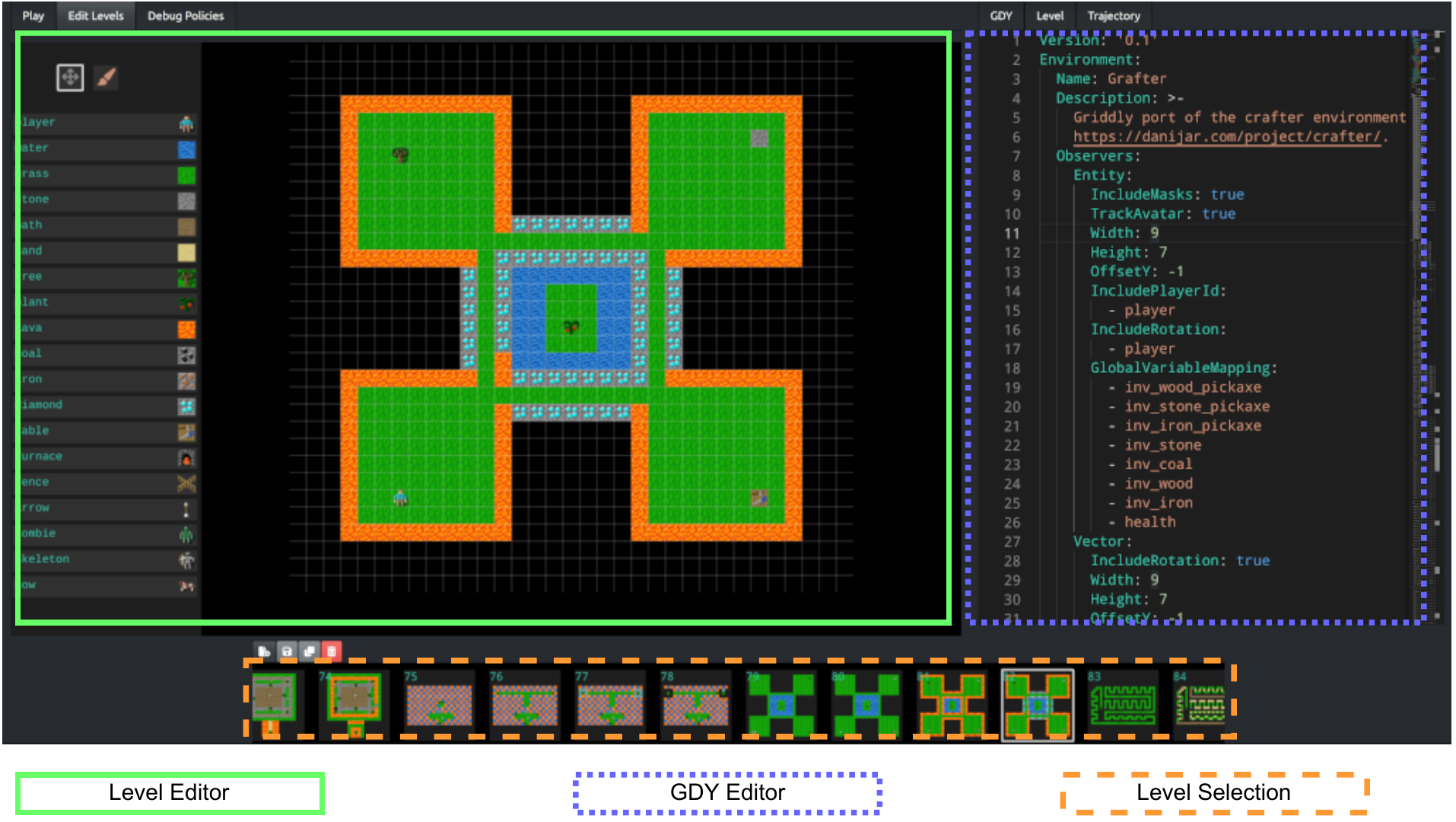}
    \caption{\small{GriddlyJS has three main components: The level editor allows rapid design of custom levels with a code-free, visual interface; rendered levels are fully interactive via keyboard control. The GDY editor allows editing of the underlying GDY specification of the core environment mechanics. The level selection component lists previously designed levels. Users can select levels for further modification or deletion.}}
    \label{fig:griddly_js}
\end{figure}

GriddlyJS provides a fully web-based integrated development environment (see Figure~\ref{fig:griddly_js}) composed of  simple and intuitive user interfaces wrapping the core components of the Griddly engine. As such, the GriddlyJS IDE can be used inside any modern browser, without the need for installing complex dependencies. Running Griddly directly inside of the browser is made possible through transcompiling the core components of Griddly into WebAssembly (Wasm). The interface itself is written using the React library. Inside the GriddlyJS IDE, GDY files can be directly edited with any changes to the environment's mechanics immediately reflected. Moreover, specific levels of the environment can be designed using a simple visual editor, allowing levels to be drawn directly inside the IDE. Previously designed levels can be saved locally into a gallery and instantly reloaded into the environment instance running inside the IDE and played via keyboard controls. Taken together, the features of GriddlyJS allow for rapid environment development, debugging, and experimentation. We now discuss the major highlights of GriddlyJS in turn. For a detailed walkthrough of these features, see Appendix \ref{sec:griddlyjs_walkthrough}.
    
\textbf{Environment Specification}\; 
Designing new environment dynamics can be time-consuming. For example, adding and testing a new reward transition requires recompiling the environment and adding specific test cases. With GriddlyJS changes can be coded directly inside the GDY in the browser, where it will be immediately reflected in the environment. The designer can then interactively control the agent to test the new dynamic. Moreover, the environment's action space is automatically reanalyzed on all changes and environment actions are assigned sensible key combinations, e.g. WASD for movement actions. Similarly, newly defined entities inside the GDY are immediately reflected in the visual level editor, allowing for rapid experimentation.

\textbf{Level Design}\; Given any GDY file, GriddlyJS provides a visual level editor that allows an end user to design environment levels by drawing tiles on a grid. Objects from the GDY file can be selected and placed in the grid by pointing and clicking. The level size is automatically adjusted as objects are added to the it. The corresponding level description string, which is used by the Griddly environment in Python to reset to that specific level, is automatically generated based on the character-to-entity mapping defined in the GDY. New levels can then be saved to the same GDY file and loaded inside the Python environment.

\textbf{Publish to the Web}\; As GriddlyJS is built using the React library, the environment component itself can be encapsulated inside a React web component. Moreover, GriddlyJS supports loading and running of TF.js models directly inside the IDE environment instance. Taken together, this allows publishing of Griddly environments and associated agent policies in the form of TF.js models directly to the web, as an embedded React web component. By allowing researchers to directly share interactive demos of their trained agents and environments, GriddlyJS provides a simple means to publish reproducible results, as well as research artifacts that encourage the audience to further engage with the strengths and weaknesses of the methods studied.

\textbf{Recording Human Trajectories}\; Recording human trajectories for environments is as simple as pressing a record button in the GriddlyJS interface and controlling the agent via the keyboard. Recorded trajectories are saved as JSON, consisting of a random seed for deterministically resetting the environment to a specific level and the list of actions taken. They can easily be compiled to datasets, e.g. for offline RL or imitation learning. The recorded trajectories can also be replayed inside of GriddlyJS. 

\textbf{Policy Visualization}\; Visualizing policy behavior is a crucial debugging technique in RL. Policy models can be loaded into GriddlyJS using TF.js and run in real-time on any level. In this way, the strengths, weaknesses, and unexpected behaviors of a trained policy can be quickly identified, providing intuition and clues about any bugs or aspects of the environment that may be challenging in a closed-loop development cycle. These insights can then be used to produce new levels that can bridge the generalization gap and thus improve the robustness of the agent.

\section{Proof-of-Concept: Escape Room Puzzles}
\label{sec:poc}

We now demonstrate the utility of GriddlyJS by rapidly creating a complex, procedurally-generated RL environment from scratch. After developing this new environment, we then use GriddlyJS to quickly hand-design a large, diverse collection of custom levels, as well as record a dataset of expert trajectories on these levels, which can be used for offline RL. We then load this new environment into Griddly for Python to train an RL agent on domain randomized levels---whose generation rules are defined within the associated GDY specification---and evaluate the agent's performance on the human-designed levels.

\subsection{Rapid Environment Development}

\begin{wrapfigure}{r}{0.33\linewidth}
\begin{minipage}[h]{\linewidth}
\vspace{-4mm}
\begin{minted}[
    gobble=2,
    frame=single,
    fontsize=\tiny, 
    breaklines
  ]{yaml}
  - Name: do
    ...
    Behaviours:
      ...
      - Src:
          Object: player
          Preconditions:
            ...
          Commands:
            - add:
                - inv_wood
                - 1
            - if:
                Conditions:
                  lt:
                    - ach_collect_wood
                    - 1
                OnTrue:
                  - set:
                      - ach_collect_wood
                      - 1
                  - reward: 1
        Dst:
          Object: tree
          Commands:
            - remove: true
            - spawn: grass    
\end{minted}
\end{minipage}
\caption{\small{GDY for an environment transition for picking up wood.}}
\label{fig:example_gdy_action}
\vspace{-4mm}
\end{wrapfigure}

We consider an environment, resembling a 2D version of MineCraft, in which the agent must learn a set of skills related to gathering resources and constructing entities using these resources in order to reach a goal. While prior works have presented environments with similar 2D, compositional reasoning challenges \citep{andreas2017modular, zambaldi2018relational, zhong2020rtfm}, we specifically model our \texttt{EscapeRoom}~ environment after the complex state and transition dynamics of Crafter \citep{hafnerBenchmarkingSpectrumAgent2021}, with the key difference being that \texttt{EscapeRoom} episodes terminate and provide a large reward upon reaching the goal object, a cherry tree, in each level.\footnote{A full description of how \texttt{EscapeRooms}~ deviates from Crafter is provided in Appendix~\ref{sec:experimental_details}.}  The dynamics inherited from Crafter entail harvesting raw resources such as wood and coal in order to build tools like furnaces, bridges, and pickaxes required to harvest or otherwise clear the path of additional resource tiles like iron and diamond.


Mastering this environment presents a difficult exploration problem for the agent: Not only must the agent reach a potentially faraway goal, but it must also learn several subskills required to reliably survive and construct a path leading to this goal. Success in this environment thus requires exploration, learning modular subskills, as well as generalization across PCG levels. Meanwhile, implementing these rich dynamics presents a time-consuming challenge for the researcher. Such a complex environment typically entails knowledge of many disparate modules performing functions ranging from GPU-accelerated graphics rendering and vectorized processes for parallelized experience collection. Further, the researcher must implement complex logic for executing the finite-state automata underlying the environment transitions, as well as that handling the rendering of observations.

\begin{figure}[t!]
\centering
\begin{subfigure}{.18\textwidth}
  \centering
  \includegraphics[height=25mm]{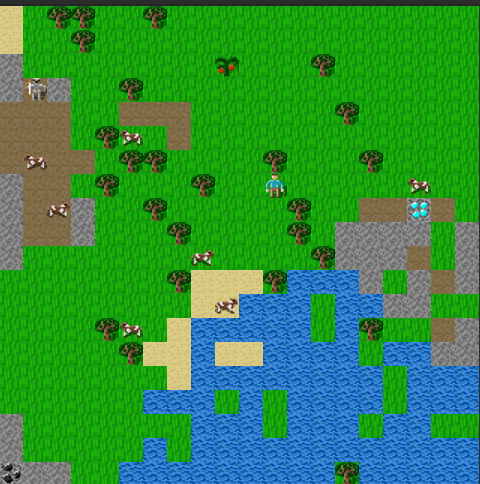}
  \caption{}
  \label{fig:crafter_generated1}
\end{subfigure}%
\hspace{0.5mm}
\begin{subfigure}{.18\textwidth}
  \centering
  \includegraphics[height=25mm]{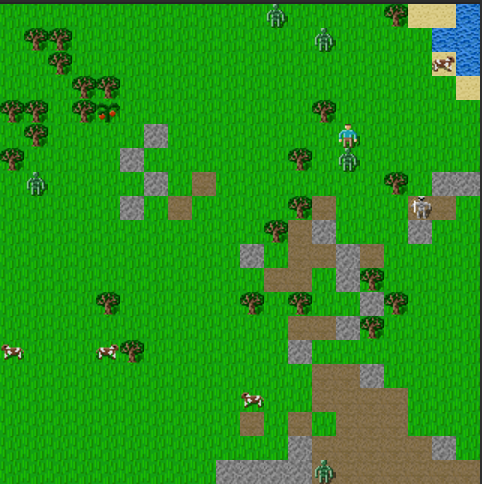}
  \caption{}
  \label{fig:crafter_generated2}
\end{subfigure}
\hspace{0.5mm}
\begin{subfigure}{.26\textwidth}
  \centering
  \includegraphics[height=25mm]{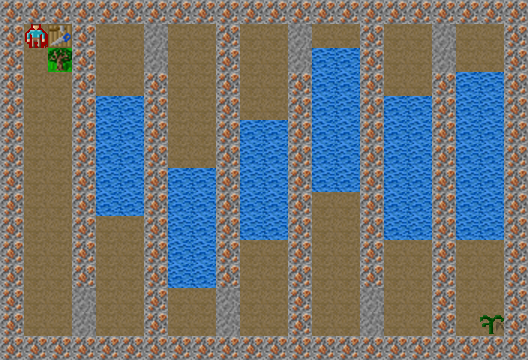}
  \caption{}
  \label{fig:level_30}
\end{subfigure}%
\hspace{0.5mm}
\begin{subfigure}{.26\textwidth}
  \centering
  \includegraphics[height=25mm]{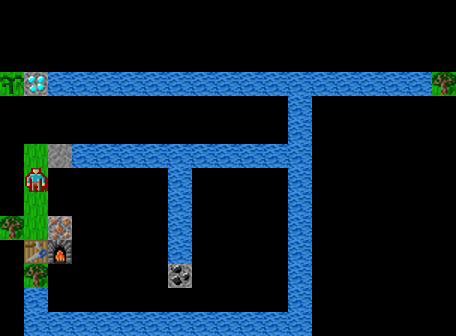}
  \caption{}
  \label{fig:level_38}
\end{subfigure}%
\caption{\small{
Example \texttt{EscapeRoom} that are procedurally-generated (a, b) and human-designed (c,d). The agent must collect resources to build tools and structures to reach the goal cherry tree, while surviving the environment.}}
\label{fig:example_escape_rooms}
\vspace{-4mm}
\end{figure}

GriddlyJS abstracts away all of these details, allowing the researcher to focus exclusively on defining the underlying MDP through a succinct GDY specification. In particular, the researcher can simply define all entities (i.e. Griddly \texttt{Objects}) present in the game, each with an array of internal state variables, as well as the agent's possible actions. Then, all transition dynamics are simply established by declaring a series of local transition rules (i.e. Griddly actions) based on the state of each entity in each tile, as well as any destination tile, acted upon by the agent's action. For example, after declaring the action of \texttt{do} (i.e. interact with an object) along with the possible game entities and their states (e.g. the agent is the \texttt{player}, which can be \texttt{sleeping}) we can simply define the transition dynamic of receiving +1 wood resource upon performing \texttt{do} on a \texttt{tree} using the simple sub-block declaration shown in Figure \ref{fig:example_gdy_action}. More complex dynamics can be implemented by calling built-in algorithms like A* search or nesting Griddly \texttt{Action} definitions. Further, arbitrary PCG logic can be easily implemented by writing Python subroutines that output level strings corresponding to the ordering of the level tiles.

\subsection{Human-in-the-Loop Level Design}
\label{subsec:hilp_design}
Given the rich design space of the \texttt{EscapeRoom} environment, randomized PCG rules defined by the Griddly level generator are unlikely to create challenging levels that push the boundaries of the agent's current capabilities. Rather in practice, designing such challenging levels for such puzzle games rely heavily on human creativity, intuition, and expertise, which can quickly hone in on the subsets of levels posing unique difficulties for a player or AI. Indeed, based on recent works investigating the out-of-distribution (OOD) robustness of RL agents trained on domain-randomized (DR) levels \citep{dennisEmergentComplexityZeroshot2021, jiangReplayGuidedAdversarialEnvironment2021}, we do not expect agents trained purely on randomized PCG levels to perform well on highly out-of-distribution, human-designed levels, without the usage of such adaptive curricula. However, it can be costly to collect a large set of diverse and challenging human-designed levels necessary to encompass the relevant challenges, and thus, most prior works test on limited sets of human-designed OOD levels. 

\begin{wrapfigure}{r}{0.48\textwidth}
    \vspace{-8mm}
    \includegraphics[width=0.9\linewidth]{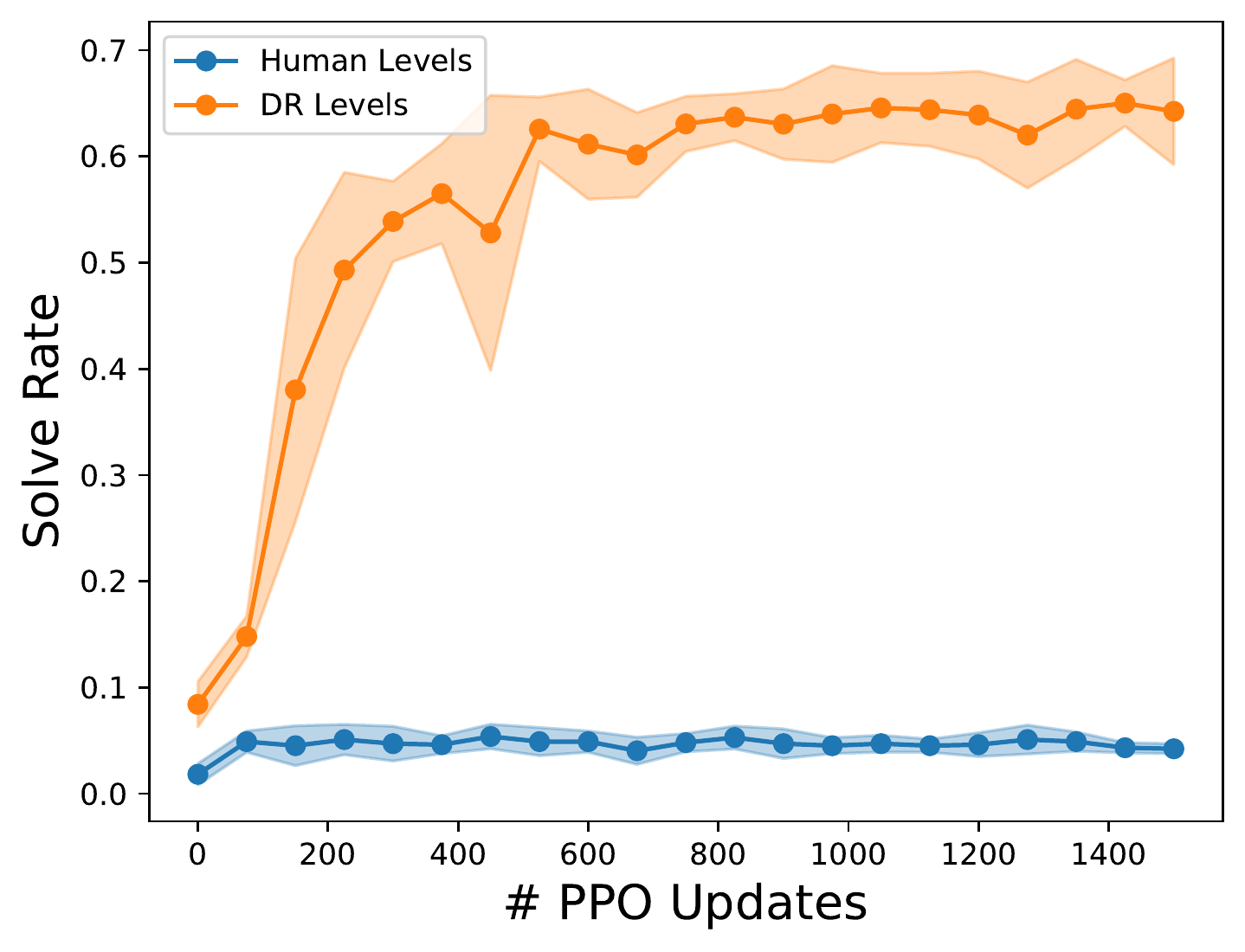} 
    \caption{\small{Mean and std of solve rate on DR levels (orange) and human-designed levels (blue).}}
    \label{fig:results}
    \vspace{-5mm}
\end{wrapfigure}

GriddlyJS allows us to quickly assess OOD generalization on a large number of human-designed levels. With its visual level editor and interactive, browser-based control of agents, we can rapidly design and iterate on new and challenging levels. In particular, we created 100 diverse environments in roughly eight hours, many featuring environment and solution structures that are highly unlikely to be generated at random. 
We then use PPO to train a policy on domain-randomized \texttt{EscapeRoom} levels. We checkpoint the policy at regular intervals in terms of number of PPO updates and evaluate the performance of each checkpoint on all 100 human-designed levels, as well as 100 DR levels. We see in Figure \ref{fig:results} that, throughout training, the resulting policy solves DR levels at a significantly higher rate than human-designed levels, highlighting the distinct quality of human-designed levels. We also note here that the human-designed levels are exponentially unlikely to fall inside the distribution of the generated levels. This highlights a significant limitation of PCG, which does not exist with levels generated by a human. Generating complex puzzle levels using PCG methods is an ongoing area of research. Tuning generators to create unique levels often results in levels that are invalid or unsolvable, and vice versa \citep{summervilleProceduralContentGeneration2018, zakariaProceduralLevelGeneration2022, siperPathDestructionLearning2022, earleIlluminatingDiverseNeural2022a, dennisEmergentComplexityZeroshot2021}. The full details on our choice of  model architecture and hyperparameters is provided in Appendix \ref{sec:experimental_details}.

Furthermore, as GriddlyJS loads TF.js models for policy evaluation and visualization directly inside the IDE, human-in-the-loop level design can be performed in a closed-loop, adversarial manner: The policy, first trained on DR levels, is successively evaluated on additional sets of human-designed levels, the most challenging of which are added to the agent's training set, thereby robustifying the agent's weaknesses.  Given the success of these methods and that of recent adversarial adaptive curricula methods for RL \citep{jiang2021prioritized} in producing robust models, we expect human-in-the-loop adversarial training to lead to similarly significant gains in policy robustness. Importantly, by developing an RL environment in GriddlyJS, this mode of training is immediately made available to the researcher. GriddlyJS enables TF.js policies and environments to be directly published on the web, thus allowing such adversarial methods to be tested at high scale, potentially leading to highly robust policies and collecting unique datasets of adversarial levels useful to future research in generalization and the emerging field of unsupervised environment design \citep{dennisEmergentComplexityZeroshot2021}.


\subsection{Recording and Controlling Trajectories}

GriddlyJS makes it easy to record trajectories for any level directly inside the IDE, enabling a wide range of downstream use cases. For example, such recorded trajectories can be associated with human-designed levels during human-in-the-loop adversarial training to ensure solvability. Further, such recorded trajectories naturally serve as datasets for offline RL and imitation learning---especially useful for more complex multi-task or goal-conditioned environments where it may be important to ensure the dataset has sufficient coverage of the various tasks or goals \cite{rudder-align}. 

Moreover, the interactive control feature in the presence of a loaded TF.js policy enables the study of human-AI interaction, in which the agent may hand over control to a human when the policy is uncertain. Further, as levels can be edited directly inside the IDE, GriddlyJS allows researchers to perform controlled evaluation of policy adaptation to environment changes that occur mid-episode.

\section{Related Works}
Several systems for developing custom procedurally-generated environments have been introduced, taking the form of code libraries featuring a domain-specific language (DSL) and encompassing both grid worlds \citep{schaulVideoGameDescription2013, perezliebana_2018, johansen2019video, beattieDeepMindLab2D2020a}, including the Griddly library for Python \citep{bamford_2021}, as well as more complex, physics-based environments \citep{scorsoglio2021visualenv, juliani2018unity}. More recently, MiniHack \citep{samvelyanMiniHackPlanetSandbox2021} builds on top of the NetHack Learning Environment \citep[NLE;][]{nle} to provide a library for developing grid worlds in the NetHack runtime. These systems focus exclusively on programmatic environment creation and provide no additional functionality for design, debugging, and agent evaluation.

Other related systems include the ML-Agents module, which converts Unity games into RL environments \citep{juliani2018unity}. The resulting trained agents can then be loaded back into Unity for evaluation. However, ML-Agents provides no specific tooling for constructing environments.
Recent works have also provided interactive evaluation interfaces, similar to that provided by GriddlyJS. TeachMyAgent \citep{romac2021teachmyagent, germon2021demo}  provides a user interface for visualizing the performance of pre-trained RL agents in a limited set of 2D continuous-control tasks extending those in OpenAI Gym \citep{brockman2016openai}. Other recent works have provided interactive, web-based demonstrations of pre-trained agents in PCG environments \citep{parkerholderEvolvingCurriculaRegretBased2022, reda2022learning}. Unlike GriddlyJS, they focus on interactive visualization, rather than providing a streamlined, closed-loop workflow connecting environment development to agent training, evaluation, and publication via the web.
Such experiments in publishing continue a rich lineage of works exploring interactive articles \citep{hohman2020communicating}, a mode of publication whose popularity GriddlyJS seeks to catalyze.

PCG has enabled many findings on RL robustness and generalization, including the important, preliminary result that RL methods easily overfit to singleton environments \citep{farebrother2018generalization, packer2018assessing, zhang2018study}. Since then, many methods for improving robustness and generalization of RL agents have been studied, notably various forms of policy regularization, data augmentation strategies \citep{raileanu2021automatic, wang2020improving}, model architecture \citep{loynd2020working,bertoin2022local}, feature distillation \citep{igl2021transient,cobbe2021phasic,raileanu2021decoupling}, and adaptive curricula \citep{portelas2020teacher,jiang2021prioritized,jiangReplayGuidedAdversarialEnvironment2021,dennisEmergentComplexityZeroshot2021,parkerholderEvolvingCurriculaRegretBased2022}. Still, the number of available PCG environments is ultimately limited to a mostly arbitrary selection determined by the interests of the select few research groups who have invested the resources to produce these environments. As such, this fast-growing area of RL research risks overfitting to a small subset of environments. By streamlining PCG environment creation, agent evaluation, and interactive result sharing, GriddlyJS seeks to empower a the wider research community.

\section{Conclusions and Future Work}
\label{sec:conclusion}
We introduced GriddlyJS, a fully web-based integrated development environment that streamlines the development of grid-world reinforcement learning (RL) environments.
By abstracting away the implementation layers responsible for shared business logic across environments and providing a visual interface allowing researchers to quickly prototype environments and evaluate trained policies in the form of TensorFlow.js models, we believe GriddlyJS can greatly improve research productivity. Moreover, in future work, we plan to use web-based tracking of anonymized user behaviors (e.g. those correlated with notions of productivity, like number of environments created and mean time for environment creation) and standard A/B testing methods to better understand how GriddlyJS is used, track bugs, and surface usability pain-points. Such a data-driven approach allows us to iteratively improving GriddlyJS for users based on direct quantifications of user productivity. 
GriddlyJS enables human-in-the-loop environment development, which we believe will become a major paradigm in RL research and development, allowing for the measured design of higher quality environments and therefore training data. Moreover, such approaches (and therefore GriddlyJS) can enable new training regimes for RL, such as human-in-the-loop adversarial curriculum learning.

Of course, our system is not without limitations: GriddlyJS does not currently persist user-generated data on a dedicated server, though we plan to support this functionality in the future. Additionally, although environments are rendered in the browser, pixel-based observation states are not currently supported. Moreover, training must still occur outside of GriddlyJS, a bottleneck that is mitigated by the fact that GDY files can be so easily loaded into Griddly for Python, and most model formats are easily converted to the TF.js format as highlighted in Appendix \ref{app:evaluating_models}.

As a proof-of-concept, we used GriddlyJS to rapidly develop the \texttt{EscapeRoom} environment based on the complex skill-based dynamics of Crafter, along with 100 custom hand-designed levels. We then demonstrated that an agent trained on domain-randomized levels performs poorly on human-designed ones. This result shows that PCG has difficulty generating useful structures for learning behaviors that generalize to OOD human-designed levels, thereby highlighting the value of GriddlyJS's simple interface for quickly designing custom levels.
Additionally, we believe GriddlyJS's web-first approach will enable more RL researchers to share their results in the form of interactive agent-environment demos embedded in a webpage, thereby centering their reporting on rich and engaging research artifacts that directly reproduce their findings.
Taken together, we believe the features of GriddlyJS have the potential to greatly improve the productivity and reproducibility of RL research.

\section{Acknowledgements}

Firstly, we would like to thank Meta for funding this research and the AI Creative Design team for the "Griddly Bear" logo. Secondly, we would like to thank the following people for testing and feedback on GriddlyJS: Remo Sasso, Sebastian Berns, Elena Petrovskaya and Akbir Khan.
Finally, we are grateful to anonymous reviewers for their recommendations that helped improve the paper.

\bibliographystyle{plainnat}
\bibliography{bibliography}

\newpage
\appendix

\section{Griddly GDY Overview} \label{sec:gdy_overview}

The Game Description YAML (GDY) is the domain-specific language of the Griddly framework that provides a means to easily design rich and diverse grid-based environments. Based on the Yet Another Markup Language (YAML), this human-readable language features simple declarative rules. GDY files are composed of three main sections, namely \texttt{Environment}, \texttt{Actions}, and \texttt{Objects}. 

To provide an intuition on how environments can be implemented using Griddly, we provide a brief tutorial of the GDY below which recreates the popular game of Sokoban. 
Once the GDY is completed, Griddly's environment wrappers can easily map it to a Gym environment.
 

\subsection{Objects}\label{sec:gdy_objects}

In this section we describe the objects of the game and the ways they can be rendered in Griddly.

The game of Sokoban is composed of four objects: \textbf{avatar}, \textbf{box}, \textbf{hole}, and \textbf{wall}.
\textbf{avatar} is the main decision-making object which can move around and push boxes into holes. Walls are immovable objects. The goal of the \textbf{avatar} is to push all boxes into the holes.\footnote{There are many variations of the game of Sokoban. In our particular implementation the agent can push boxes into any hole on the environment. There also exist versions where only a single box can be pushed into each hole.}

Each object needs to have a unique name, which can serve as references to that object in other parts of the GDY code. Let us start the \texttt{\textcolor{darkgreen}{Objects}} section and define the \textbf{avatar} object as follows:

\begin{lstlisting}[frame=none]
Objects:
 - Name: avatar
   Z: 2
   MapCharacter: A
   Observers:
     Sprite2D:
       Image: images/gvgai/oryx/knight1.png
\end{lstlisting}

\texttt{\textcolor{darkgreen}{MapCharacter}} is used to define the ASCII character of an object and can be used to mark the initial positions of objects in concrete levels defined later in the \texttt{\textcolor{darkgreen}{Environment}} section.

The property \textit{\textcolor{darkgreen}{Z}} can serve as the third dimension of the cells in the grid. It allows to define objects that can occupy the same location of the grid, as long as they have different \textit{\textcolor{darkgreen}{Z}} values. It also defines the order of objects when rendered in Griddly, i.e., higher \textit{\textcolor{darkgreen}{Z}} values indicate that the objects will be rendered on top.

The \texttt{\textcolor{darkgreen}{Observers}} property determines how each observer type will render this particular object. Here, the \textbf{avatar} object only includes a Sprite2D observer, but Griddly supports additional forms of observations, including those shown in Figure \ref{fig:observers}. Sprite2D observers expect an image for rendering, thus we select a knight icon from Griddly's selection of icons.\footnote{Griddly allows users to easily upload new custom icons for their own environments.}

Having finished the description of the \textbf{avatar} object, we proceed to the \textbf{wall} object. Here we provide 16 different images for walls to correspond to different positions of walls, such as horizontal or vertical locations, corner pieces, T-pieces, etc.

\begin{lstlisting}[frame=none]
- Name: wall
  MapCharacter: w
  Observers:
    Sprite2D:
      TilingMode: WALL_16
      Image:
        - images/gvgai/oryx/wall3_0.png
        - images/gvgai/oryx/wall3_1.png
        - images/gvgai/oryx/wall3_2.png
        - images/gvgai/oryx/wall3_3.png
        - images/gvgai/oryx/wall3_4.png
        - images/gvgai/oryx/wall3_5.png
        - images/gvgai/oryx/wall3_6.png
        - images/gvgai/oryx/wall3_7.png
        - images/gvgai/oryx/wall3_8.png
        - images/gvgai/oryx/wall3_9.png
        - images/gvgai/oryx/wall3_10.png
        - images/gvgai/oryx/wall3_11.png
        - images/gvgai/oryx/wall3_12.png
        - images/gvgai/oryx/wall3_13.png
        - images/gvgai/oryx/wall3_14.png
        - images/gvgai/oryx/wall3_15.png
\end{lstlisting}

We do not provide this object with a \texttt{\textcolor{darkgreen}{Z}} value given that nothing should interact with this object. WALL\_16 tiling mode is used to make sure all 16 wall icons are rendered correctly for each location.

\begin{lstlisting}[frame=none]
- Name: box
  Z: 2
  MapCharacter: b
  Observers:
    Sprite2D:
      Image: images/gvgai/newset/block1.png

- Name: hole
  Z: 1
  MapCharacter: h
  Observers:
    Sprite2D:
      Image: images/gvgai/oryx/cspell4.png
\end{lstlisting}

The \textbf{box} and \textbf{hole} objects can be defined  similar to the \textbf{avatar} objects, except that the \textbf{hole} objects have a different \textcolor{darkgreen}{Z} value allowing the avatar to move on top of them.

\subsection{Actions}\label{sec:gdy_actions}

\texttt{\textcolor{darkgreen}{Actions}} define the mechanics of the game and interactions between objects in Griddly. Each individual action includes two entities: \textbf{source} and \textbf{destination}. \textbf{source} is the object which performs a particular action, whilst the \textbf{destination} is the object that is affected by this action. Firstly, we define the movement action of the \textbf{avatar} as follows.

\begin{lstlisting}[frame=none]
Actions:
 # Define the move action
 - Name: move
   Behaviours:
   # The agent can move around freely in empty space and over holes
     - Src:
         Object: avatar
         Commands:
           - mov: _dest
       Dst:
         Object: _empty
\end{lstlisting}

Given that the  \texttt{\textcolor{darkgreen}{Src}} key includes \textbf{avatar} object as its value, it can be inferred that this is an action performed by the \textbf{avatar}. The \texttt{\textcolor{darkgreen}{Dst}} key with the object value \_empty indicates that the behaviour only applies when the action is performed on a space with no objects on it.

The \texttt{\textcolor{darkgreen}{Commands}} property in the \texttt{\textcolor{darkgreen}{Src}} field includes a list of instructions that will be executed to the \texttt{\textcolor{darkgreen}{Src}} object once this action is performed. The \texttt{\textcolor{darkgreen}{mov}:~\_dest} commands moves the object to the destination of the action. 

Next, we define the box pushing actions. Firstly, we define the ability of \textbf{box} objects to move to empty locations. Then, we allow the \textbf{avatar} object to interact with the \textbf{box} object.

\begin{lstlisting}[frame=none]
# Boxes can move into empty space
- Src:
    Object: box
    Commands:
        - mov: _dest
  Dst:
    Object: _empty

# The agent can push boxes
- Src:
    Object: avatar
    Commands:
        - mov: _dest
  Dst:
    Object: box
    Commands:
        - cascade: _dest
\end{lstlisting}

Here we make sure that the \textbf{box} object is moved in the same direction as the \textbf{avatar} object, the source of the action. We achieve this by using the \texttt{\textcolor{darkgreen}{cascade:}~\_dest} which re-apply the same action on the destination object, namely the \textbf{box}.

Finally, we define the mechanics of the \textbf{box} being pushed onto a \textbf{hole}. We achieve this by defining our last action with \textbf{box} as its source and \textbf{hole} as its destination:

\begin{lstlisting}[frame=none]
# If a box is moved into a hole remove it
 - Src:
     Object: box
     Commands:
       - remove: true
       - reward: 1
   Dst:
     Object: hole
\end{lstlisting}

Here, the \texttt{\textcolor{darkgreen}{remove:}~true} command removes the source \textbf{box} from the grid once pushed into a hole. Furthermore, the \texttt{\textcolor{darkgreen}{reward:}~1} commands Griddly to provide the agent with the reward of 1 once this event is triggered.

\subsection{Environment}\label{sec:gdy_env}

The \texttt{\textcolor{darkgreen}{Environment}} section defines the environment description, such as its name, as well as the observation and action spaces of the MDP. For all environments, Griddly uses the canonical action space layout from \cite{bamfordGeneralisingDiscreteAction2021a}, with support for invalid action masking. In Griddly, \textit{observers} are functions that determine how any given environment state is transformed into an observation. Several different types of observers are supported, including those shown in Figure \ref{fig:observers}. Griddly additionally supports lightweight observers based on default 2D block sprites, ASCII, and semantic entity states (as defined in the GDY). Observers are highly customizable, with options for partial observability, rotation, cropping, and inclusion of global or internal, entity-level variables. 

Below we provide the \texttt{\textcolor{darkgreen}{Environment}} section for our Sokoban example. 
Firstly, we indicate that the \textbf{avatar} object will serve as the decision-making agent in our environment:

\begin{lstlisting}[frame=none]
Player:
  AvatarObject: avatar
\end{lstlisting}

We then describe the termination condition which determines when the episode is complete and whether the agent wins or loses. For Sokoban, an episodes is considered won if all the boxes are pushed into the holes, i.e., the number of boxes in the environment is equal to 0:

\begin{lstlisting}[frame=none]
Termination:
    Win:
     - eq: [box:count, 0]
\end{lstlisting}

Our next step is to defines levels for our Sokoban game. The layout of each level can be defined using a sequence of string that uses the \texttt{\textcolor{darkgreen}{MapCharacter}} characters of each object defined above. The dot character \texttt{.} indicates an unoccupied grid cell.

\begin{lstlisting}[frame=none]
Levels:
    - |
      wwwwwww
      w..hA.w
      w.whw.w
      w...b.w
      whbb.ww
      w..wwww
      wwwwwww
    - |
      wwwwwwwww
      ww.h....w
      ww...bA.w
      w....w..w
      wwwbw...w
      www...w.w
      wwwh....w
      wwwwwwwww
\end{lstlisting}

\begin{figure}
    \centering
    \includegraphics[width=0.22\linewidth]{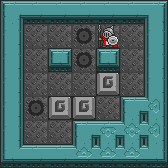}
    \includegraphics[width=0.3\linewidth]{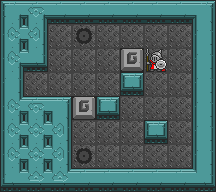}
    \caption{Custom Sokoban levels defined in the GDY example.}
    \label{fig:example_levels}
\end{figure}

The two defined levels produce the environment renderings illustrated in Figure \ref{fig:example_levels}.

Lastly, we specify the size of the tiles in pixels and the background image using the \texttt{\textcolor{darkgreen}{MapCharacter}} and \texttt{\textcolor{darkgreen}{BackgroundTile}} fields. We also provide the environment with a unique name.

\begin{lstlisting}[frame=none]
Environment:
   Name: sokoban
   TileSize: 24
   BackgroundTile: images/gvgai/newset/floor2.png
\end{lstlisting}

\subsection{Putting It All Together}

Figures \ref{fig:gdy_sokoban_1} and \ref{fig:gdy_sokoban_2} provide the full implementation of the Sokoban example in Griddly.

\begin{figure}[H]
\begin{minted}[
    gobble=0,
    frame=single,
    fontsize=\small, 
    breaklines
  ]{yaml}

Environment:
   Name: sokoban
   TileSize: 24
   BackgroundTile: images/gvgai/newset/floor2.png
   Player:
     AvatarObject: avatar
   Termination:
     Win:
       - eq: [box:count, 0]
   Levels:
     - |
       wwwwwww
       w..hA.w
       w.whw.w
       w...b.w
       whbb.ww
       w..wwww
       wwwwwww
     - |
       wwwwwwwww
       ww.h....w
       ww...bA.w
       w....w..w
       wwwbw...w
       www...w.w
       wwwh....w
       wwwwwwwww

Actions:
- Name: move
  Behaviours:
    - Src:
        Object: avatar
        Commands:
          - mov: _dest
      Dst:
        Object: [_empty, hole]

    - Src:
        Object: box
        Commands:
            - mov: _dest
      Dst:
        Object: _empty

    - Src:
        Object: avatar
        Commands:
          - mov: _dest
      Dst:
        Object: box
        Commands:
          - cascade: _dest

    - Src:
        Object: box
        Commands:
          - remove: true
          - reward: 1
      Dst:
        Object: hole
\end{minted}
\caption{Full implementation of Sokoban in Griddly (part 1).\label{fig:gdy_sokoban_1}}
\end{figure}

\begin{figure}[H]
\begin{minted}[
    gobble=0,
    frame=single,
    fontsize=\small, 
    breaklines
  ]{yaml}
  
Objects:
 - Name: box
   Z: 2
   MapCharacter: b
   Observers:
     Sprite2D:
       Image: images/gvgai/newset/block1.png

 - Name: wall
   MapCharacter: w
   Observers:
   Sprite2D:
     TilingMode: WALL_16
     Image:
       - images/gvgai/oryx/wall3_0.png
       - images/gvgai/oryx/wall3_1.png
       - images/gvgai/oryx/wall3_2.png
       - images/gvgai/oryx/wall3_3.png
       - images/gvgai/oryx/wall3_4.png
       - images/gvgai/oryx/wall3_5.png
       - images/gvgai/oryx/wall3_6.png
       - images/gvgai/oryx/wall3_7.png
       - images/gvgai/oryx/wall3_8.png
       - images/gvgai/oryx/wall3_9.png
       - images/gvgai/oryx/wall3_10.png
       - images/gvgai/oryx/wall3_11.png
       - images/gvgai/oryx/wall3_12.png
       - images/gvgai/oryx/wall3_13.png
       - images/gvgai/oryx/wall3_14.png
       - images/gvgai/oryx/wall3_15.png

 - Name: hole
   Z: 1
   MapCharacter: h
   Observers:
     Sprite2D:
       Image: images/gvgai/oryx/cspell4.png

 - Name: avatar
   Z: 2
   MapCharacter: A
   Observers:
     Sprite2D:
       Image: images/gvgai/oryx/knight1.png
\end{minted}
\caption{Full implementation of Sokoban in Griddly (part 2).\label{fig:gdy_sokoban_2}}
\end{figure}

\section{GriddlyJS UI Walkthrough} \label{sec:griddlyjs_walkthrough}

In this section we show various screenshots of the GriddlyJS user interface and highlight various useful features.

\subsection{Building And Testing Environment Mechanics}

\begin{figure}[h]
  \centering
  \includegraphics[width=0.9\textwidth]{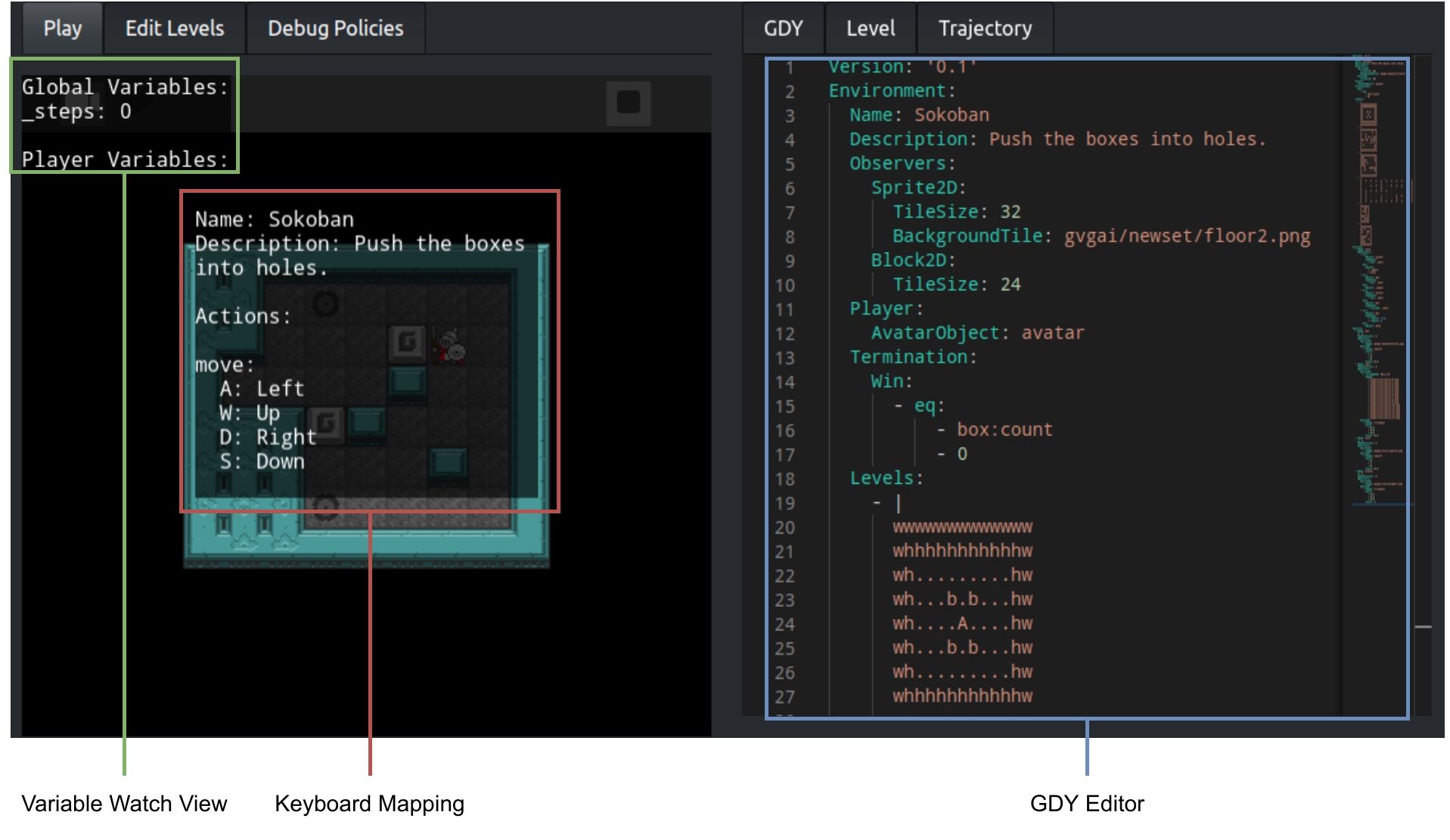}  
  \caption{\small{Environment Designing and Debugging interfaces.}.} 
  \label{fig:building_and_testing}
\end{figure}

GriddlyJS provides many tools for building and debugging environments. Figure \ref{fig:building_and_testing} shows several of these. Firstly, as soon as  GDY file is loaded in the editor and a level selected, it will be playable in the editor window. Actions in the environment are automatically mapped to the keyboard, and an explanation of this mapping can be toggled by pressing \textbf{P}. Additionally all global and player-wise variables can be toggled by pressing \textbf{I}. These variables are updated live while the game is being played. 

\subsection{Level Design}

\begin{figure}[h]
  \centering
  \includegraphics[width=0.9\textwidth]{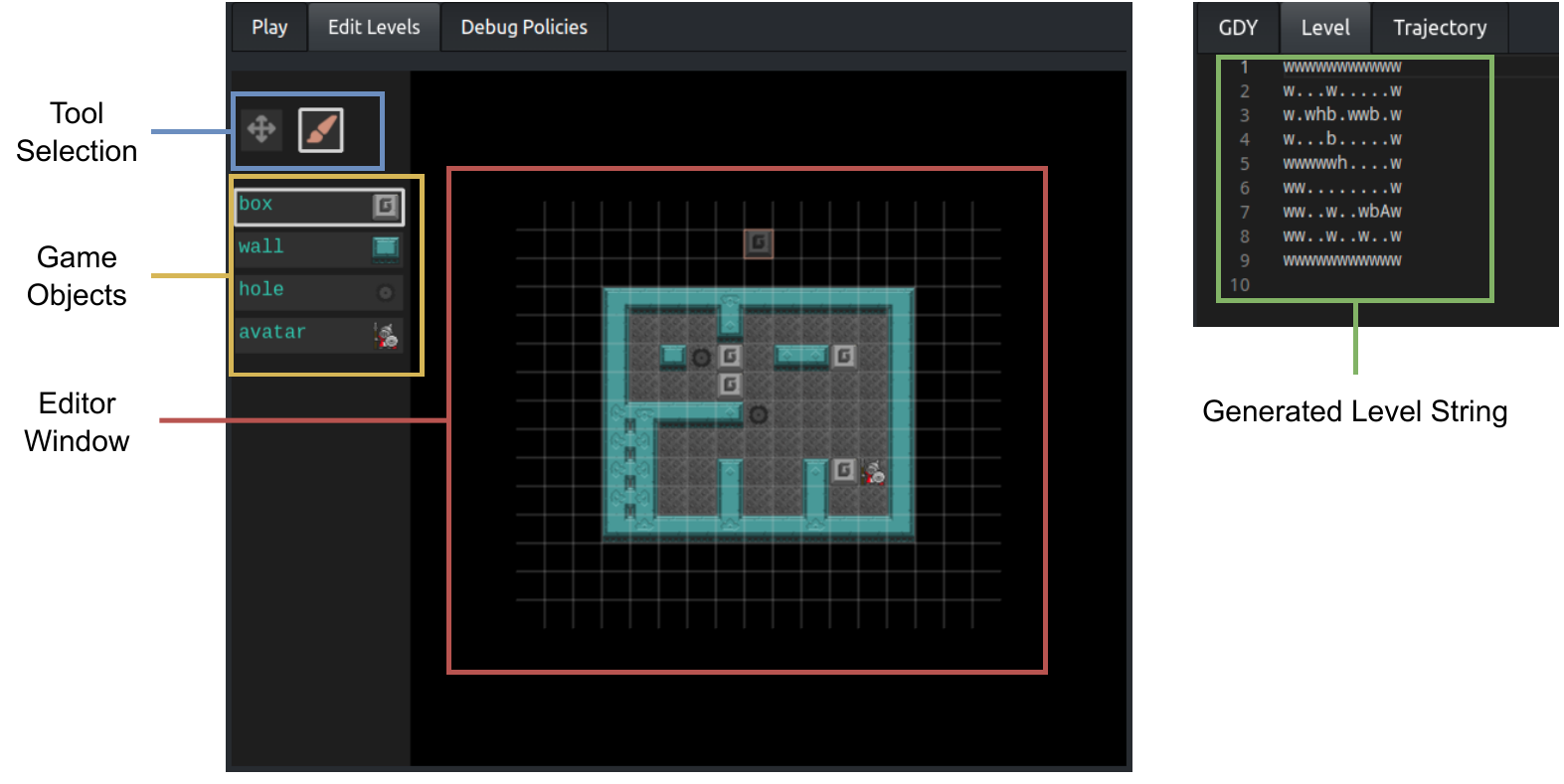}  
  \caption{\small{Level Editor view and level string view.}.} 
  \label{fig:level_editor_view}
\end{figure}

The Level Editor view shown in Figure \ref{fig:level_editor_view} allows the user to choose objects to place on the game grid in order to create levels. The user can selects an object from the menu and then can {\em paint} it into the editing grid. The editing grid automatically grows if objects are placed near the boundaries, so levels of any shape or size can be created. Additionally as objects are painted into the editor grid, a level string is automatically generated and displayed. This level string can also be manually edited and its changes reflected in the editor window. 

\begin{figure}[h]
  \centering
  \includegraphics[width=0.9\textwidth]{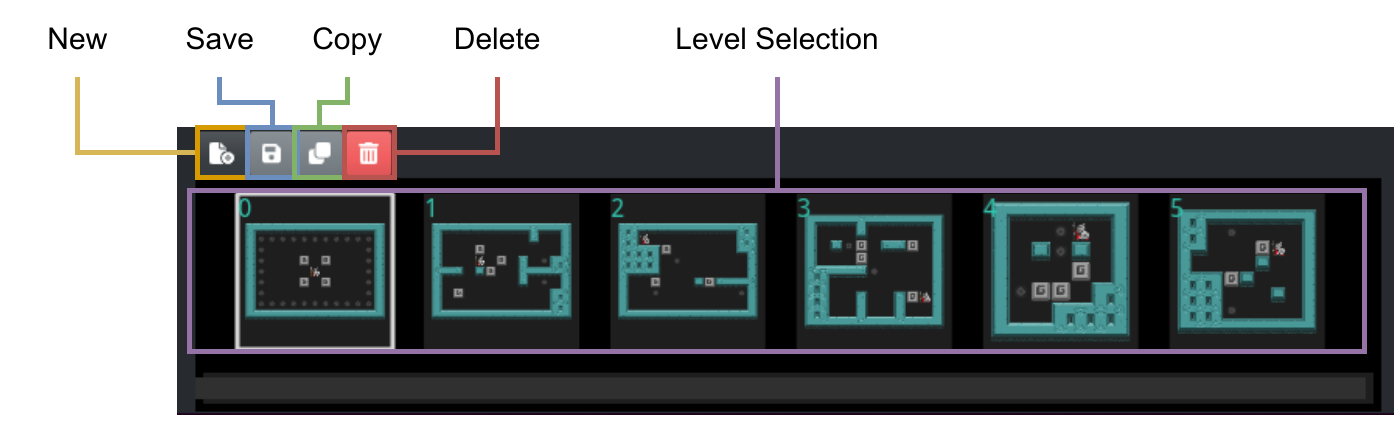}  
  \caption{\small{The level selector view showing thumbnails of the the levels in the GDY.}.} 
  \label{fig:level_selection}
\end{figure}

Managing the set of levels in the environments GDY file is handled by the Level Selection interface shown in Figure \ref{fig:level_selection}. Users can create, update and delete levels in the GDY file for quickly generating large datasets of levels.

\subsection{Recording Trajectories}

\begin{figure}[h]
  \centering
  \includegraphics[width=0.9\textwidth]{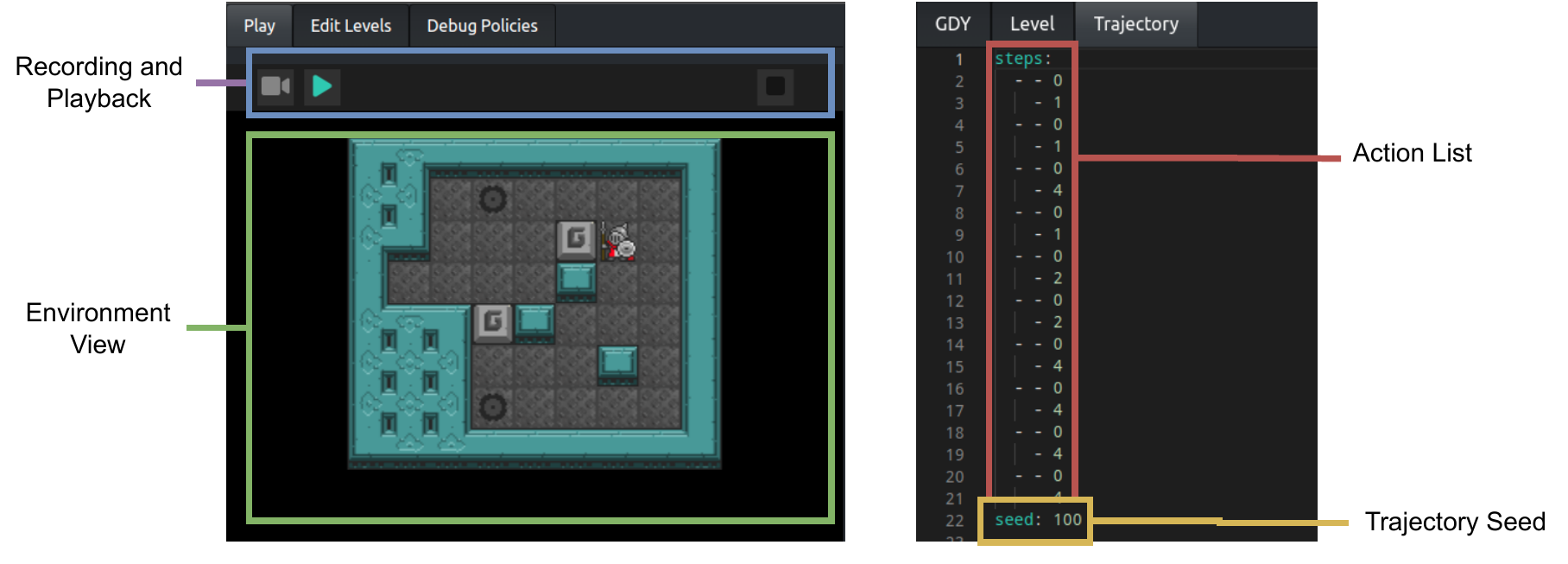}  
  \caption{\small{Recording and playback menus for generating  and viewing trajectories}.} 
  \label{fig:recording_and_playback}
\end{figure}

Recording and playback options in the GriddlyJS interface. On clicking the \textbf{Record} icon, the user's actions are recorded as they play the game in the environment view. When the environment terminates, it is reset to its original state and a \textbf{play} button is shown next to the record button. If pressed, the play button re-plays the recorded trajectory. These options are shown in Figure \ref{fig:recording_and_playback}. Additionally, the actions and seed are displayed as YAML in the trajectory view in the editor. This trajectory can then be copied and stored for later use, for example in behavioural cloning algorithms. Trajectories can also be copied into the text-area from external sources and played within the editor.

\subsection{Evaluating Models} \label{app:evaluating_models}

\begin{figure}[h]
  \centering
  \includegraphics[width=0.5\textwidth]{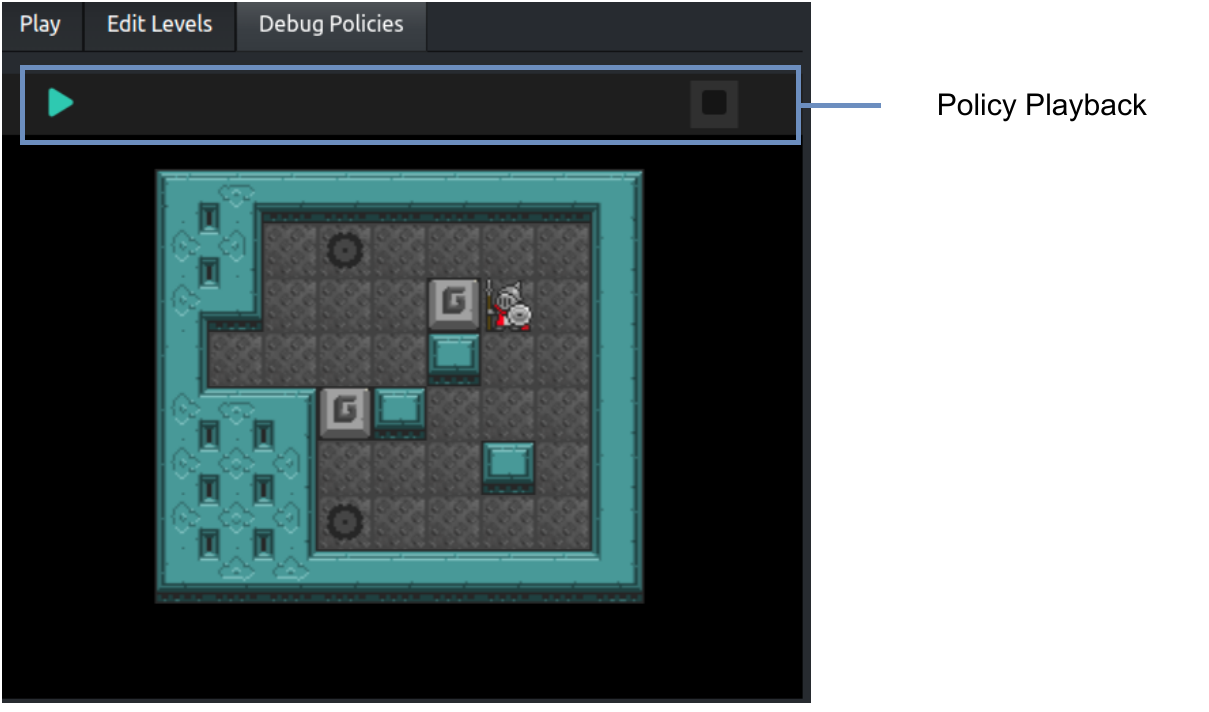}  
  \caption{\small{Policy debugger view, similar to the "play" view but only visible if a policy is loaded using TensorflowJS.}} 
  \label{fig:policy_debugger_view}
\end{figure}

Trained policies can be loaded into GriddlyJS and replayed using the Debug Policies view, shown in Figure \ref{fig:policy_debugger_view}. If a model is loaded, a \textbf{play} button will be visible which will sample actions from the policy to view its performance. once the episode is finished, the level is reset.

\subsubsection{Deep Learning Framework Support}

GriddlyJS supports any deep learning model that can be converted into the ONNX format~\citep{ONNXHome}. This includes many popular frameworks  such as \textbf{PyTorch}, \textbf{Tensorflow}, \textbf{JAX}, \textbf{Caffe} and \textbf{Chainer}. Once converted to the ONNX format, these models can be converted to TensorflowJS and used in the debugging view.

As our experiments in section \ref{sec:poc} are trained using PyTorch, we include example scripts to convert these models to ONNX and then to TensorflowJS. These scripts and documentation on how to load and use the converted models can be found at: \url{https://github.com/GriddlyAI/escape-rooms#using-checkpoints-in-griddlyjs}

\section{Experimental Details and Hyperparameters}\label{sec:experimental_details}

Table \ref{table:hyperparam_sweeps} summarises the hyperparameters we chose to sweep. Other hyperparameters while sweeping were those shown in Table \ref{table:hyperparams}. 

Table \ref{table:hyperparams} summarises our final hyperparameter choices for our PPO agent.The final choice was made by taking highest average (calculated across the seeds) level completion rate.

\begin{table}[h!]
\caption{Hyperparameter sweep values}
\label{table:hyperparam_sweeps}
\begin{center}

\begin{tabular}{lrr}
\toprule
\textbf{Parameter} & Values  \\
\midrule
$\lambda_{\text{GAE}}$ & 0.65, 0.8, 0.95  \\
Adam learning rate & 5e-2, 1e-2, 5e-3, 1e-3, 5e-4, 1e-4  \\
Student entropy coefficient & 0.2, 0.1, 5e-2, 1e-2, 5e-3, 1e-3  \\
Seeds & 0 1 2 3 4 5 6 7 8 9 \\
\bottomrule 
\end{tabular}
\end{center}
\end{table}

\begin{table}[h!]
\caption{Hyperparameters used for training the PPO model.}
\label{table:hyperparams}
\begin{center}

\begin{tabular}{lrr}
\toprule
\textbf{Parameter} & Values  \\
\midrule
$\gamma$ & 0.99  \\
$\lambda_{\text{GAE}}$ & 0.95  \\
PPO rollout length & 128  \\
PPO epochs & 4  \\
PPO minibatches per epoch & 4 \\
PPO clip range & 0.2  \\
PPO number of workers & 256 \\
Adam learning rate & 1e-3  \\
Adam $\epsilon$ & 1e-5 \\
PPO max gradient norm & 0.5 \\
PPO value clipping & yes \\
Return normalization & no  \\
Value loss coefficient & 0.5  \\
Student entropy coefficient & 0.05  \\

\bottomrule 
\end{tabular}
\end{center}
\end{table}

\subsection{Architecture}

We use the PPO implementation from CleanRL \citep{huang2021cleanrl} with the {\em ImpalaCNN} \citep{espeholt_2018} architecture as this is commonly used with grid-world environments. 

\subsection{Training And Evaluation}

All training and evaluation episodes are limited to 500 steps. The agent receives no penalty for reaching this limit. We trained our models with 10 different seeds for 50 million environment steps. All training is performed using our modified Crafter level generator as described in the next section. All results are averaged across these 10 seeds. All code for our experiments and descriptions on how to use the training and evaluation scripts can be found in the escape-rooms repository: \url{https://github.com/GriddlyAI/escape-rooms}

\subsection{Modified Crafter Environment}
\label{subsec:modified_crafter}

Griddly's GDY format contains and restricted set of commands that allow complex mechanics to be realised. However when translating from many environments into GDY format, there are some caveats that may mean behaviours are slightly modified from the original versions. 

\subsubsection{Grafter}

Grafter Github repository: \url{https://github.com/GriddlyAI/grafter}

Before generating the Escape Room environments, Crafter was first translated directly to GDY to create as close a replication of the original environment as possible. This replication (Nicknamed Grafter) had several features that could not be directly translated. These translation artifacts between the environment implementations are explained below:

\textbf{Chunk Balancing}\; The spawning and despawning of Non-player characters (NPC) i.e zombies, cows, and skeletons in order to balance their numbers across the environment is not possible using the current features of Griddly, so objects are only spawned at the start of the episode. Defeating NPCs removes them permanently from the environment.

\textbf{Day and Night Cycles}\; Changing the brightness of pixels in order to simulate a day and night cycle is relevant only when pixel observations are being used. Griddly supports several other observation spaces where day and night cannot be easily modelled, such as \texttt{Vector} and \texttt{Entity}. Day and night is still implemented as part of the \texttt{Sprite2D} observations, but the associated behavioural changes for NPCs are not present. 

\textbf{Chasing Behaviour}\; Zombies and skeletons use Griddly's built-in A* pathfinding implementation, whereas zombies and skeletons in Crafter use a simple rule-based method.

\textbf{Observation Spaces}\; All observation spaces are configured to contain the same information as the original crafter environment and have equivalent observability dimensions. 
\textbf{Sprite2D} observers configured to be the same as the original Crafter environment, the inventory display and day/night cycle are produced by a custom shader. In this implementation, the voronai pixel noise used in the original environment is ommitted.
\textbf{Vector} observers contain a 7x9x51 (WxHxC) observation space, where the channels $C$ represent object types, orientations, playerIds and the set of global variables which represent the inventory, which are repeated across the height and width dimensions.
\textbf{Entity} observer contain a list of features for each object type in the 7x9 space around the agent and additionally include a {\em global} entity which contains the inventory variables.

\paragraph{Multi-Agent Support} is naturally introduced as part of the features that come with Griddly, agents gain an additional achievement if they defeat other agents. The number of agents that are spawned in the environment is configurable in the GDY.

\subsubsection{Domain Randomization}

\begin{figure}[ht!]
    \centering
    \includegraphics[width=\textwidth]{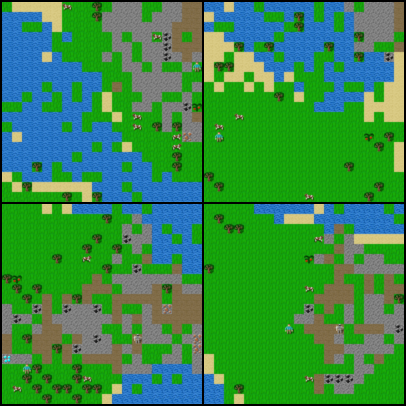}
    \caption{\small{Example escape rooms generated by the Domain Randomization generator.}}
    \label{fig:dr-levels}
\end{figure}

The domain randomization algorithm we use is a modified version of the open-simplex based level generator from the original Crafter environment. While the structure of the levels is generally the same as those in Crafter, we also make sure that we add a single "cherry" tree goal to each level. In most cases the cherry tree can be reached by traversing land, but occasionally there may be levels where more complex strategies are required such as chopping trees or building bridges to get to the island where a tree exists. We show examples of the DR levels in Figure \ref{fig:dr-levels}

\subsubsection{Escape Rooms}

Escape Rooms Github repository \url{https://github.com/GriddlyAI/escape-rooms}

\begin{table}[h!]
\caption{Flattened Action Space}

\label{table:actions}
\begin{center}

\begin{tabular}{lrr}
\toprule
\textbf{Action} & Values  \\
\midrule
No-Op & 0 \\
Move Left & 1  \\
Move Right & 2  \\
Move Down & 3  \\
Move Up & 4 \\
Interact With Object & 5  \\
Place Stone & 6 \\
Place Table & 7 \\
Place Furnace & 8 \\
Make Wood Pickaxe & 9 \\
Make Stone Pickaxe & 10 \\
Make Iron Pickaxe & 11 \\

\bottomrule 
\end{tabular}
\end{center}
\end{table}

To make an escape room, firstly we needed a method of \textbf{escape}. To do this we repurposed the mechanic of eating a \textbf{plant} object. We added a termination condition so that if the {\em eat plant} goal is achieved, the episode ends and the agent receives a reward of 10. 

There are also several features in Grafter that were not required in the Escape Room environment:

\textbf{Plants}\; As reaching and eating a plant (cherry tree) is now being used as the goal state, the mechanics for collecting, planting and ripening of the trees was removed. Plants are spawned in the {\em ripe} state and remain that way until the agent collects them and subsequently ends the episode.

\textbf{Agent Survival}\; The mechanics for surviving in the environment, such as requiring food, water, energy and maintaining health levels are unnecessary complexities for the escape rooms and limit the possible challenges that can be built. The same reasoning is applied to zombies and skeletons which are not required. Removing these mechanics also removes the need for certain actions such as sleeping. Additionally, the day/night mechanics were removed entirely.

\textbf{Swords}\; Similar to the reasoning behind survival mechanics, we decided that combat with zombies/skeletons was an unnecessary complication, therefore the mechanics for building weapons were not required. This also simplified the action space as is shown in table \ref{table:actions}. 

\textbf{Reward Shaping}\; In Crafter the agent is rewarded depending on their current health level, as we are not using any health or survival mechanics this reward scheme is ommitted. All achievements still give a single reward of 1 for the first time they are encountered. The exception being the {\em eat plant} achievement which gives a reward of 10 for completing the escape room.

\section{Solution Trajectories} \label{sec:solution_trajectories}
\begin{figure}[ht!]
  \centering
    \centering
    \includegraphics[width=\textwidth]{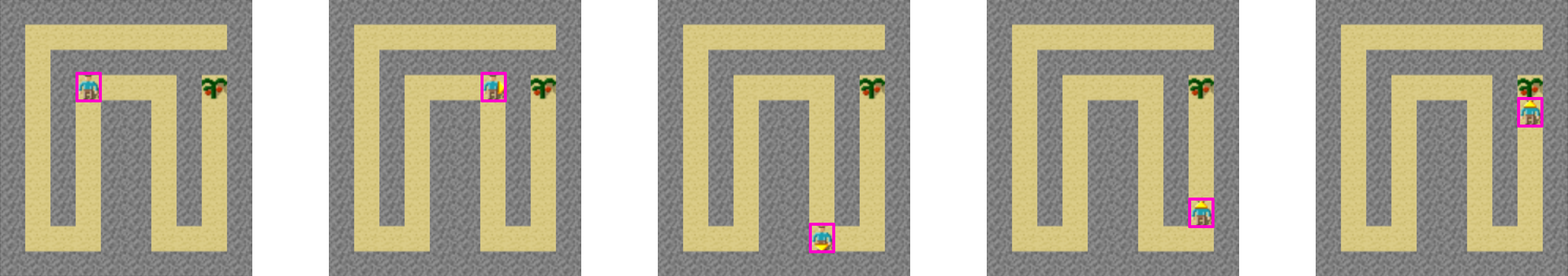}
    \caption{\small{\textbf{Level 1.} In this simple level, the agent must solely navigate the maze to reach the goal cherry tree.}}
    \label{fig:solution_trajectories_1}
    \label{fig:eg_traj_1}
\end{figure}

\begin{figure}[ht!]
    \centering
    \includegraphics[width=\textwidth]{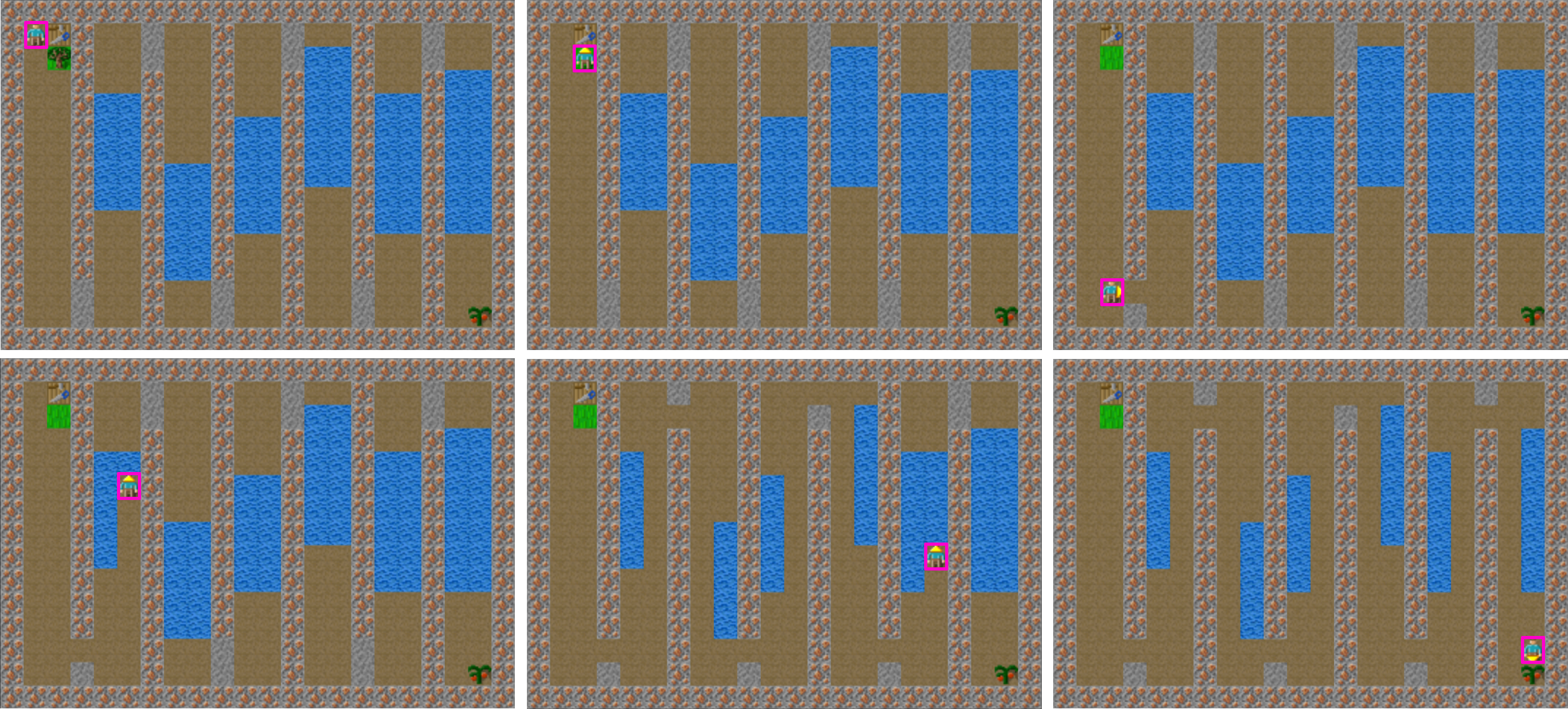}
    \caption{\small{\textbf{Level 30.} The agent begins in the top-left corner, where it must first collect wood and build a wooden pickaxe using the work table. With this tool, the agent can pick up stones to clear the path, as well as successively place and pick up stones over the water to create a walkable path, making it possible to reach the goal.}}
    \label{fig:eg_traj_2}
\end{figure}

\begin{figure}[ht!]
    \centering
    \includegraphics[width=\textwidth]{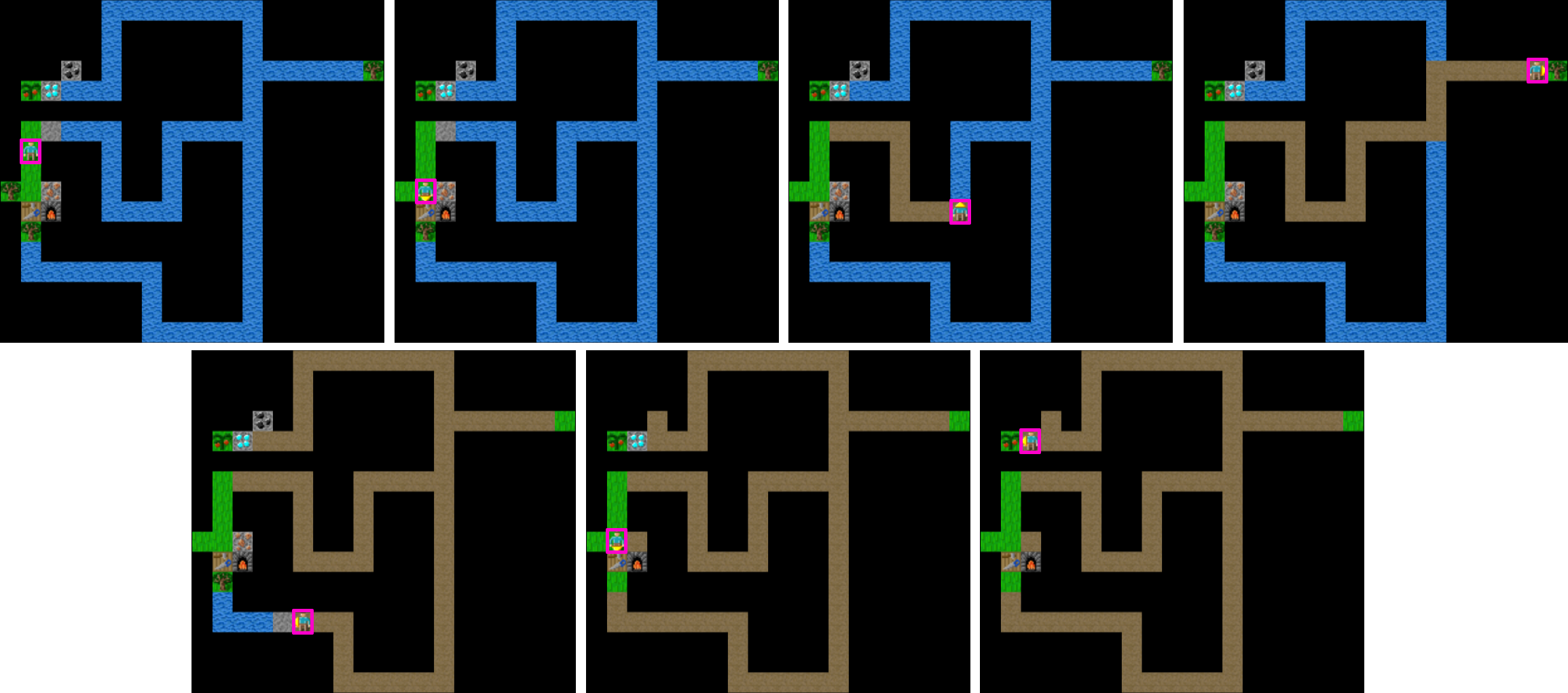}
    \caption{\small{\textbf{Level 37.} The agent starts on the left and must first collect the wood and go to the work table to build a wooden pickaxe, with which it can collect the stone above. The agent must then successively place and remove this stone over the water to make the water walkable paths, making sure to collect the remaining two pieces of wood. Returning to the work table, the agent can then build a stone pickaxe, collect the coal at the top of the level, return to the work table and furnace to build an iron pickaxe, with which it can clear the diamond blocking the goal.}}
    \label{fig:eg_traj_3}
\end{figure}

\begin{figure}[ht!]
    \centering
    \includegraphics[width=\textwidth]{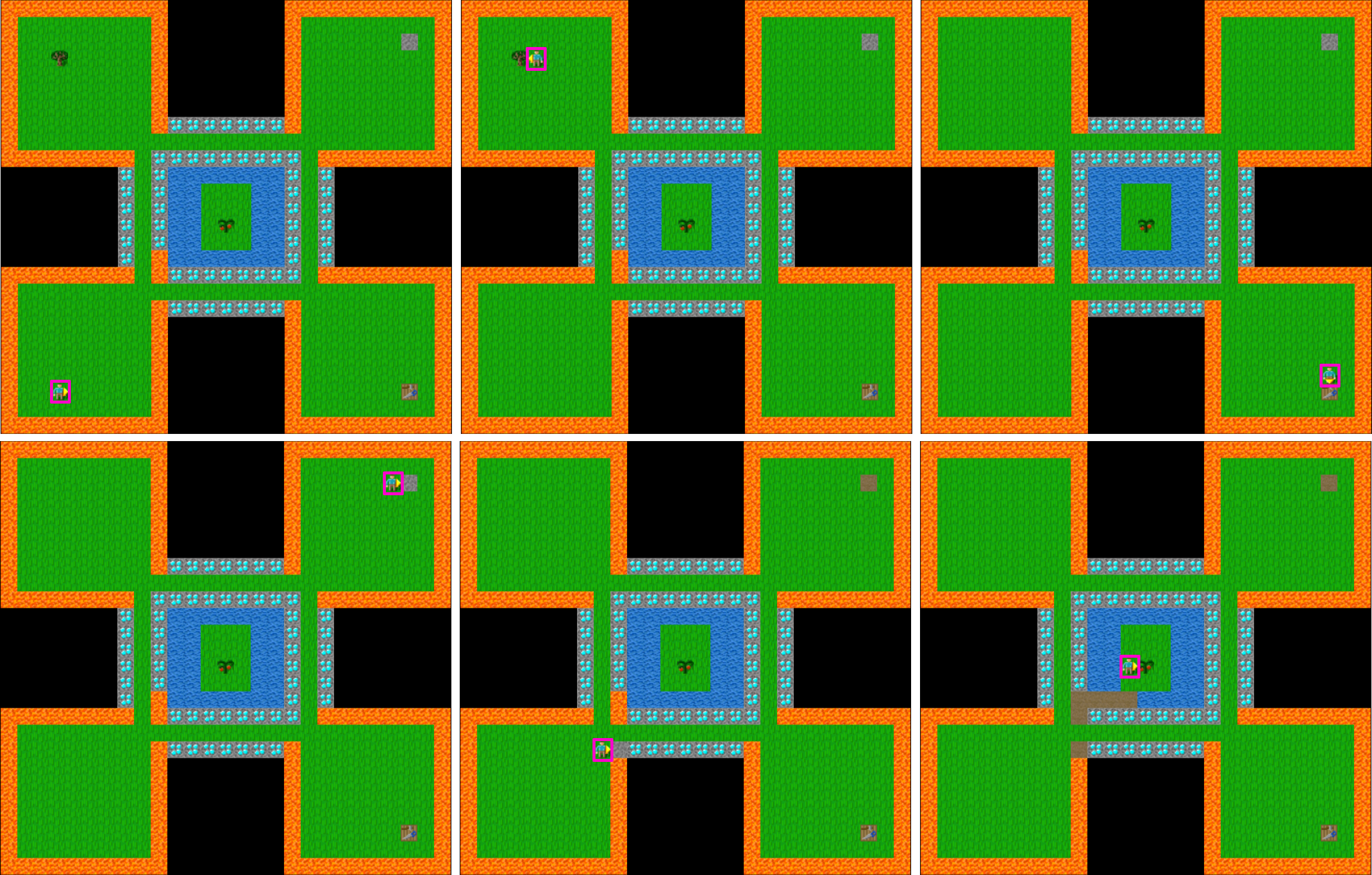}
    \caption{\small{\textbf{Level 82.} The agent begins in the bottom-left corner and then must visit top-left corner to collect wood, visit the work table in the bottom-right to create a wood pickaxe, collect the stone in the top-right corner, and then successively place and remove the stone over the lava to create a walkable path through the lava corner in at the bottom-left of the central diamond-bordered square to reach the goal.}}
    \label{fig:eg_traj_4}
\end{figure}

\begin{figure}[ht!]
    \centering
    \includegraphics[width=\textwidth]{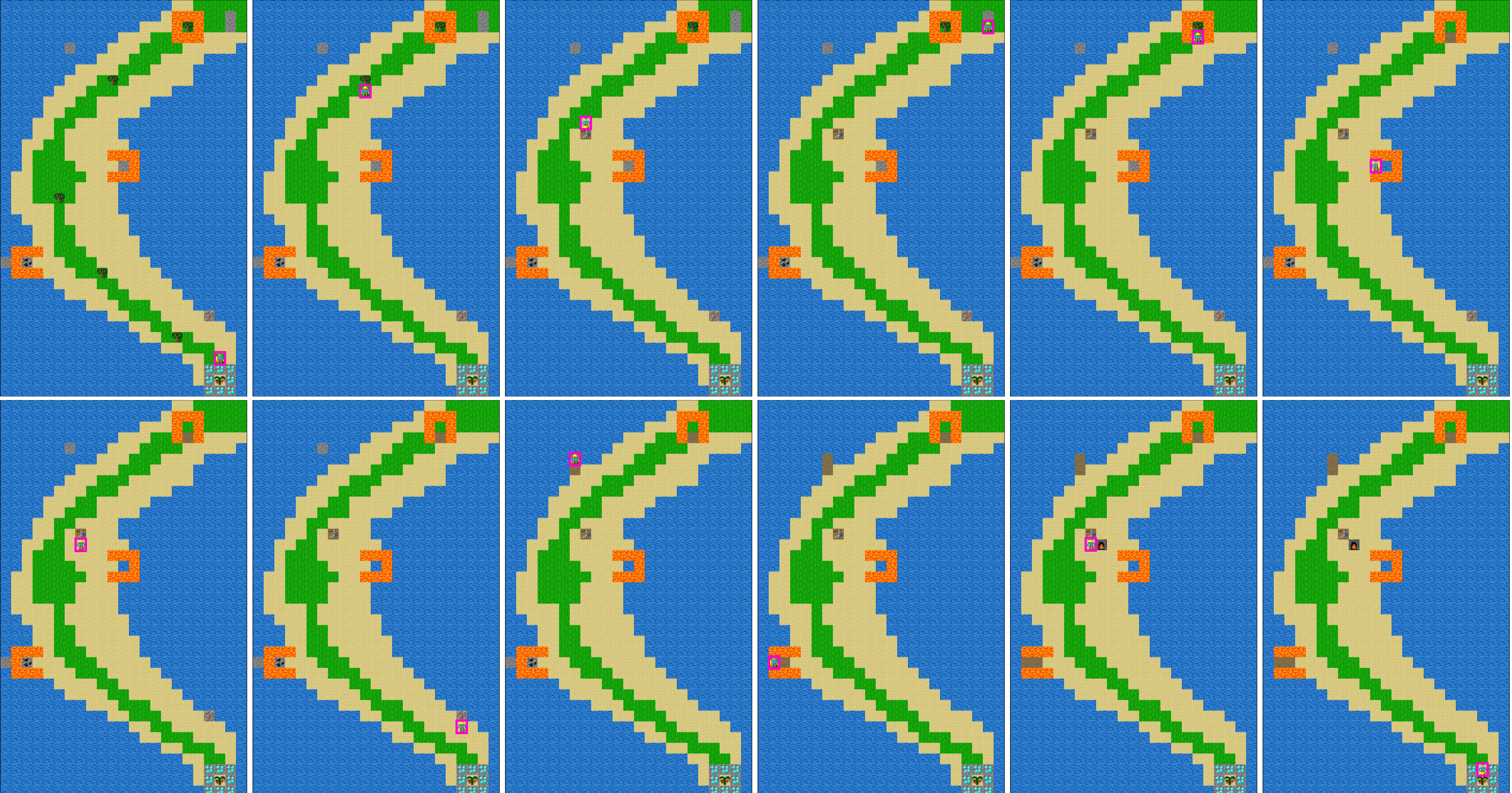}
    \caption{\small{\textbf{Level 95.} The agent begins at the bottom of the island. It must collect the wood across the island to build a work table and then a wooden pickaxe, with which it can collect stone ---which requires building a path to the stone in the water by placing and removing stone in the water. The agent must then return to the work table to build a stone pickaxe, with which it can collect the iron in the bottom-right. With the iron, the agent must return to the work table to create a stone pickaxe, which can be used to collect the coal, clearing the way to also place and remove stone over the lava to collect the final piece of stone. The agent must then return to the work table to build a furnace and then an iron pickaxe, with which it can use to clear the diamond blocking the goal.}}
    \label{fig:eg_traj_5}
\end{figure}

\begin{figure}[ht!]
    \centering
    \includegraphics[width=\textwidth]{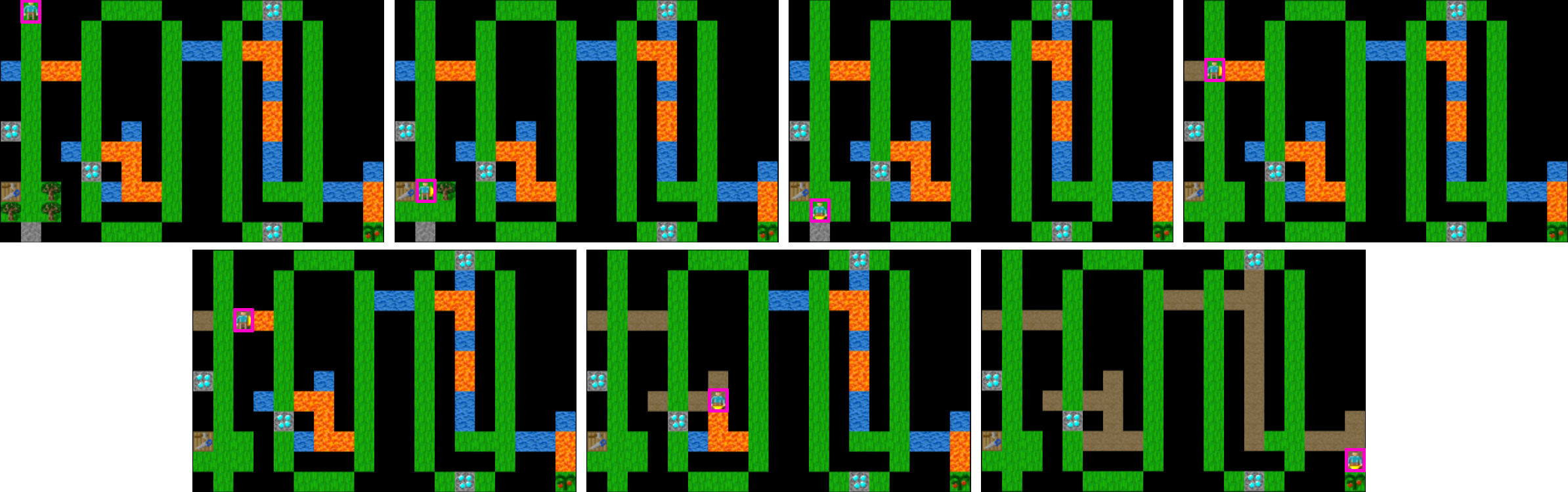}
    \caption{\small{\textbf{Level 100.} The agent starts at the top-left, and must first move down to collect the wood and build a wooden pickaxe to collect the stone. With the stone, the agent must create walkable area over each of the crevices along the path in order to allow it to properly face the lava tiles, so that the agent can then place and remove the stone over the lava to create a walkable path towards the goal. This level creates difficulty by exploiting how moving towards water does not result in episode termination, while turning into lava does.}}
    \label{fig:eg_traj_6}
\end{figure}

\begin{figure}[ht!]
    \centering
    \includegraphics[width=\textwidth]{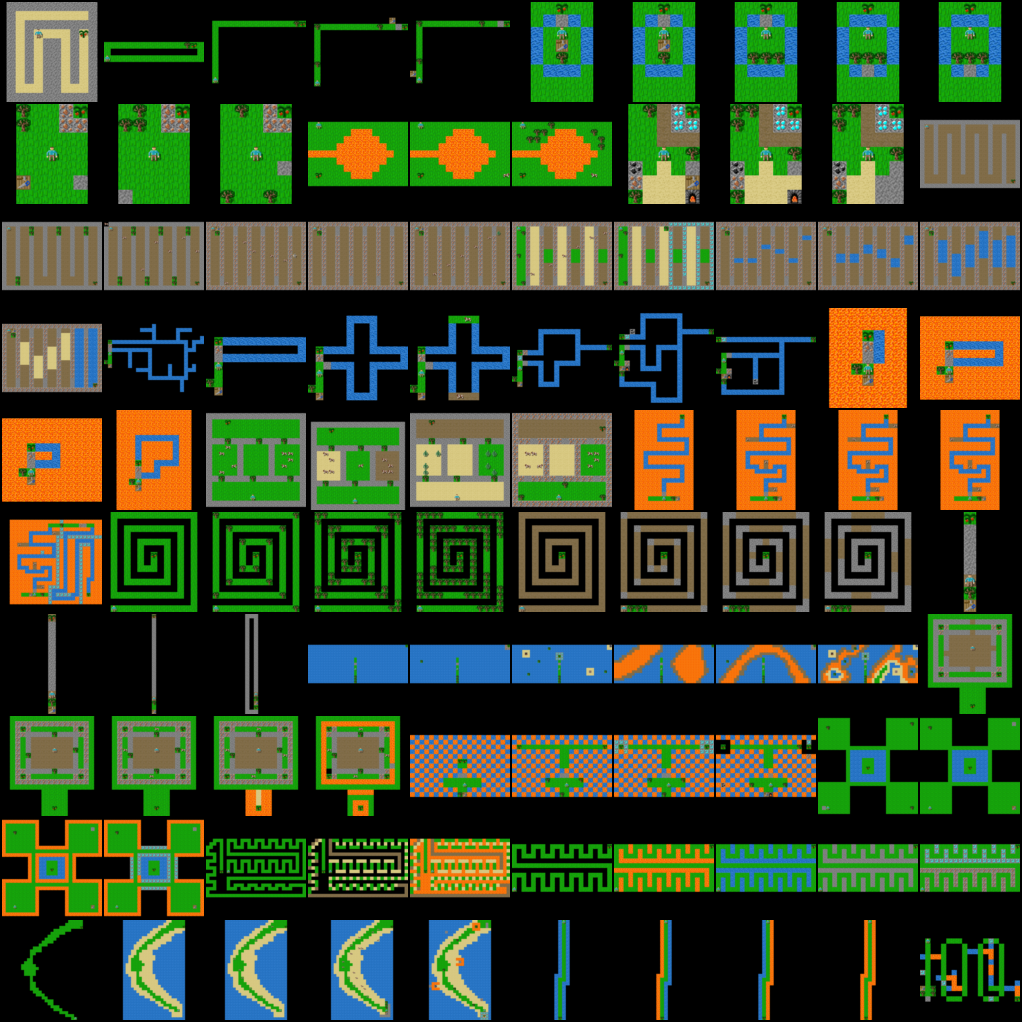}
    \caption{\small{All 100 human-designed EscapeRoom levels, made using GriddlyJS.}}
    \label{fig:100_escape_rooms}
\end{figure}

To demonstrate the ease of creating custom levels and recording trajectories with GriddlyJS IDE, we create a set of 100 hand-designed levels of the Crafter-based \texttt{EscapeRoom} environment. All 100 levels are visualized in Figure~\ref{fig:100_escape_rooms}. The levels are feature distinct challenges for the agent. For each level, we include a solution trajectory generated by a human player using the recording feature inside the IDE. Griddly stores trajectories as simply a list of actions taken, along with a string representation of the level or a specific seed that allows the level generator to deterministically reset to the recorded level. We visualize key frames (left to right, top to bottom) from expert trajectories for a diverse subset of the 100 hand-designed \texttt{EscapeRoom} levels in Figures~\ref{fig:eg_traj_1} through \ref{fig:eg_traj_6}. In each frame, the agent is highlighted with a \textcolor{magenta}{magenta} bounding-box for clarity.

\section{Assets} 
\label{sec:assets}

In this section we outline the assets and open source software used in GriddlyJS, and any associated licenses.

\textbf{Oryx Design Lab}\; As Griddly is built as a successor to GVGAI for use in Reinforcement Learning, many of the assets used in GVGAI are used in Griddly. As the original assets in GVGAI are from the Oryx Design lab\footnote{http://www.oryxdesignlab.com}, we re-purchased the equivalent asset packs for use in Griddly as per the license agreement \url{https://www.oryxdesignlab.com/license}.

\textbf{GVGAI}\; Many of the Griddly environments are inspired by those in GVGAI. For example the levels and mechanics in games like Sokoban are clones of those in GVGAI. GVGAI is distributed under the following GNU GPL license: \url{https://github.com/GAIGResearch/GVGAI/blob/master/LICENSE.txt}

\textbf{Griddly}\; Griddly uses additional asset packs from the Oryx Design Lab as well as those used in GVGAI. 
A full list of asset packs that Griddly uses:
\begin{itemize}
  \item \textbf{Iso Dungeon} \url{https://www.oryxdesignlab.com/products/iso-dungeon}
  \item  \textbf{Tiny Galaxy} \url{https://www.oryxdesignlab.com/products/tiny-galaxy-tileset}
  \item  \textbf{16-bit Fantasy} \url{https://www.oryxdesignlab.com/products/16-bit-fantasy-tileset}
\end{itemize}

As GriddlyJS is an extension on top of Griddly, we distribute GriddlyJS under the same MIT license: \url{https://github.com/Bam4d/Griddly/blob/develop/LICENSE}

\textbf{Crafter}\; As we clone the crafter environment, we also copy the assets used. Crafter and its assets are released under the MIT license: \url{https://github.com/danijar/crafter/blob/main/LICENSE}

\section{Broader Impact}
\label{sec:Broader Impact}

GriddlyJS aims to drastically improve the productivity of RL research, by streamlining the pipeline from environment development to agent training and evaluation as a closed-loop workflow, and by enabling researchers to easily publish their findings as interactive agent-environment demos that invite active inquiry from the community. By furthering progress in RL, likely a major component of the most powerful AI systems of the future, GriddlyJS aligns with AI progress. Thus, our work aligns with the downsides of more rapid AI progress as well, namely the potentially faster proliferation of autonomous systems that may be put to malicious uses, such as the spread of misinformation and the deliberate or unintentional magnification of social and economic inequalities. 

Furthermore, by simplifying the development of RL environments, the systematic biases in GriddlyJS's design may be amplified at scale, resulting in future RL research to overfit to these biases---the most obvious being GriddlyJS's deliberate, exclusive focus on grid world environments. Therefore, it is important to develop deeper understanding of the limitations of the grid world environments produced by GriddlyJS, for example, by conducting experiments comparing the computational efficiency and generalization properties of RL agents trained inside grid worlds compared to continuous control environments, under a diverse set of MDP formulations.

\end{document}